\documentclass[lettersize,journal]{IEEEtran}
\usepackage{amsmath,amsfonts}
\usepackage{algorithm}
\usepackage{array}
\usepackage[caption=false,font=normalsize,labelfont=sf,textfont=sf]{subfig}
\usepackage{textcomp}
\usepackage{stfloats}
\usepackage{url}
\usepackage{verbatim}
\usepackage{graphicx}
\usepackage{cite}

\usepackage{booktabs}
\usepackage{multirow}
\usepackage{lipsum}
\usepackage{hyperref}
\usepackage{breakurl}
\usepackage[edges]{forest}
\definecolor{hidden-draw}{RGB}{20,68,106}
\definecolor{hidden-pink}{RGB}{255,245,247}
\usepackage[framemethod=tikz]{mdframed}
\usepackage{soul}
\usepackage{times}
\usepackage{tabularx}
\usepackage{float}
\usepackage{xcolor} 
\usepackage{latexsym}
\usepackage{bbm}
\usepackage{amsfonts}
\usepackage{dsfont}
\usepackage{graphicx}
\usepackage{caption}
\usepackage{algpseudocode}
\usepackage{colortbl}
\definecolor{lightgray}{gray}{0.9}
\definecolor{lightgreen}{rgb}{0.9, 1, 0.9}
\usepackage{microtype}
\usepackage{inconsolata}
\usepackage{ulem}

\usepackage{tikz}
\usetikzlibrary{fadings}
\usetikzlibrary{mindmap,trees}
\usetikzlibrary{arrows,automata,shapes,positioning,shadows,trees}
\usetikzlibrary{shapes,snakes,shadows}
\usetikzlibrary{shapes.arrows}
\usetikzlibrary{calc,shapes, positioning}
\usetikzlibrary{decorations.text}
\usetikzlibrary{matrix,chains,positioning,decorations.pathreplacing,arrows}
\usetikzlibrary{bayesnet}
\usetikzlibrary{shadows.blur}
\usetikzlibrary{shapes.geometric}

\usepackage{pgfplots}
\usepackage{pgfplotstable}
\usepackage{pgfmath,pgffor}
\usepgfplotslibrary{colorbrewer}
\pgfplotsset{compat = 1.14, cycle list/Set1-8}
\usetikzlibrary{pgfplots.statistics, pgfplots.colorbrewer}
\pgfplotsset{compat=1.8}

\tikzstyle{edge}=[-latex',draw=black!90,shorten <=1pt,shorten >=1pt]
\tikzstyle{redge}=[latex'-,draw=black!90,shorten <=1pt,shorten >=1pt]
\tikzstyle{dedge}=[latex'-latex',draw=black!90,shorten <=1pt,shorten >=1pt]

\tikzstyle{block}=[draw, text width=5em,align=center,shape=rectangle, rounded corners, , align=center]
\tikzstyle{nobox}=[align=center]
\definecolor{emb}{RGB}{209,228,252}
\definecolor{hidden-blue}{RGB}{194,232,247}
\definecolor{hidden-orange}{RGB}{224,224,224}
\definecolor{hidden-yellow}{RGB}{242,244,193}
\definecolor{output-purple}{RGB}{219,203,231}
\definecolor{output-green}{RGB}{204,231,207}
\definecolor{hiddendraw}{RGB}{10,128,122}
\definecolor{myred2}{HTML}{F875AA}
\definecolor{mypurple2}{HTML}{D2E0FB}

\definecolor{myred}{HTML}{CD5C5C}
\definecolor{myorange}{HTML}{CD853F}
\definecolor{mypurple}{HTML}{FFDFDF}
\definecolor{myyellow}{HTML}{FFF6F6}
\definecolor{mygreen}{HTML}{2E8B57}
\definecolor{myblue}{HTML}{4682B4}
\definecolor{mygrey}{HTML}{808000}
\definecolor{myherf}{HTML}{0000CD}
\newcommand{\myherf}{\color{myherf}}
\begin{document}

\title{Graphs Meet AI Agents: Taxonomy, Progress, and Future Opportunities}

\author{
Yuanchen Bei, Weizhi Zhang, Siwen Wang, Weizhi Chen, Sheng Zhou, Hao Chen, Yong Li, Jiajun Bu, \\ Shirui Pan,~\IEEEmembership{Senior Member,~IEEE}, 
Yizhou Yu,~\IEEEmembership{Fellow,~IEEE}, Irwin King,~\IEEEmembership{Fellow,~IEEE},\\ Fakhri Karray,~\IEEEmembership{Fellow,~IEEE}, Philip S. Yu,~\IEEEmembership{Fellow,~IEEE}

\thanks{Yuanchen Bei, Weizhi Chen, Sheng Zhou, and Jiajun Bu are with Zhejiang University, Hangzhou, China. E-mail: yuanchenbei@zju.edu.cn; chenweizhi@zju.edu.cn; zhousheng\_zju@zju.edu.cn; bjj@zju.edu.cn.}% <-this % stops a space
\thanks{Weizhi Zhang and Philip S. Yu are with University of Illinois Chicago, Chicago, USA. E-mail: wzhan42@uic.edu; psyu@uic.edu.}
\thanks{Siwen Wang and Fakhri Karray are with Mohamed Bin Zayed University of Artificial Intelligence, Abu Dhabi, UAE. E-mail: wangsiwendut@gmail.com; Fakhri.Karray@mbzuai.ac.ae.}
\thanks{Hao Chen is with City University of Macau, Macao SAR, China. E-mail: sundaychenhao@gmail.com. Yong Li is with Ant Group, Hangzhou, China. E-mail: liyong.liy@antgroup.com. Shirui Pan is with Griffith University,  Brisbane, Australia. E-mail: s.pan@griffith.edu.au. Yizhou Yu is with The University of Hong Kong, Hong Kong SAR, China. E-mail: yizhouy@acm.org. Irwin King is with The Chinese University of Hong Kong, Hong Kong SAR, China. E-mail: king@cse.cuhk.edu.hk.}
}

\markboth{Preprint}%
{Shell \MakeLowercase{\textit{et al.}}: A Sample Article Using IEEEtran.cls for IEEE Journals}

\maketitle

\begin{abstract}
AI agents have experienced a paradigm shift, from early dominance by reinforcement learning (RL) to the rise of agents powered by large language models (LLMs), and now further advancing towards a synergistic fusion of RL and LLM capabilities. This progression has endowed AI agents with increasingly strong abilities. 
Despite these advances, to accomplish complex real-world tasks, agents are required to plan and execute effectively, maintain reliable memory, and coordinate smoothly with other agents. Achieving these capabilities involves contending with ever-present intricate information, operations, and interactions.
In light of this challenge, data structurization can play a promising role by transforming intricate and disorganized data into well-structured forms that agents can more effectively understand and process. In this context, graphs, with their natural advantage in organizing, managing, and harnessing intricate data relationships, present a powerful data paradigm for structurization to support the capabilities demanded by advanced AI agents.
To this end, this survey presents a first systematic review of how graphs can empower AI agents. Specifically, we explore the integration of graph techniques with core agent functionalities, highlight notable applications, and identify prospective avenues for future research. By comprehensively surveying this burgeoning intersection, we hope to inspire the development of next-generation AI agents equipped to tackle increasingly sophisticated challenges with graphs. Related resources are collected and continuously updated for the community in the \myherf{\href{https://github.com/YuanchenBei/Awesome-Graphs-Meet-Agents}{Github link}}.

\end{abstract}


\section{Introduction}\label{sec:intro}
In the rapidly evolving landscape of artificial intelligence, AI agents have received remarkable attention due to their potential to automate task processing. 
These agents have undergone a transformative journey, progressing from early reinforcement learning (RL)-based agents~\cite{nguyen2020deep,florensa2018automatic} to large language model (LLM)-powered agents~\cite{wang2024survey,zhao2024expel}, and more recently to architectures that tightly couple LLMs as the foundational knowledge backbone with RL as task-specific learning paradigms~\cite{peiyuan2024agile}.
This progression marks a substantial leap in agent capabilities, enabling them to understand complex tasks by leveraging the extensive foundational world knowledge inherent in LLMs and learning to accurately process tasks with RL optimization.

\begin{figure*}[t]
\centering
\includegraphics[width=\linewidth]{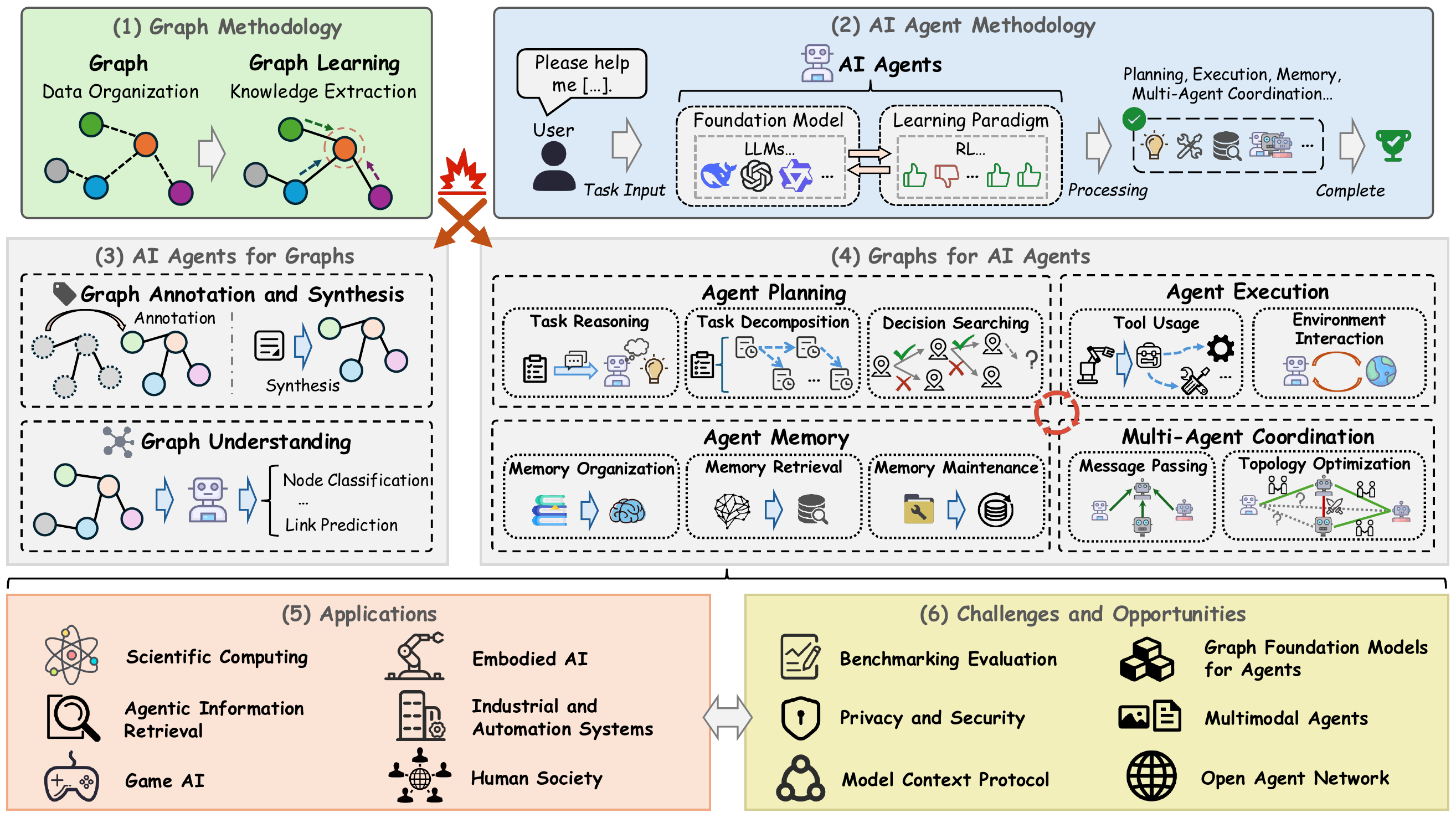}
\caption{An overall illustration of graphs meet AI agents. (1) Graph Methodology: graph data organization and knowledge extraction with graphs. (2) AI Agent Methodology: LLM-based foundation models and RL-based learning paradigms, forming core AI Agent pipelines. (3) AI Agents for Graphs: the powerful capabilities of agents in graph modeling and learning, such as graph annotation, synthesis, and understanding. (4) Graphs for AI Agents: the role and potential of graphs and graph learning in empowering the core functionalities of agents, including agent planning, execution, memory, and multi-agent coordination. (5) Representative Applications. (6) Challenges and Future Opportunities.}
\label{fig:overall}
\end{figure*}

AI agents performing complex real-world tasks are often required to possess diverse capabilities~\cite{masterman2024landscape,krishnan2025ai}. Effective task navigation requires multiple agent functionalities, such as sophisticated planning, precise execution involving external tools, maintenance of reliable memory systems, and facilitating effective coordination with other agents~\cite{huang2024understanding,guo2024large}. 
However, a central challenge is that, due to the complexity of the tasks, agents are often \textbf{\textit{confronted with intricate and disorganized information, operators, and interactions}} in these functionalities. 
It is essential to find an effective way to organize and manage such encountered data in agent functionalities so that agents can better comprehend and efficiently process the information presented to them, ultimately enhancing their effectiveness in handling complex tasks.
For example, agents need to parse unstructured task descriptions and reorganize them into an actionable schedule of subtasks during planning. The myriad of external tools available should be arranged to preserve both efficiency and accuracy for use during agent execution. 
In addition, the vast contents of the memory should be organized so that useful information is retained and readily retrievable for agents. Moreover, in a multi-agent setting, a coordination topology should be determined that enables agents to transmit meaningful messages to one another effectively.
With unstructured data, traditional agents only implicitly recognize latent correlations among information, operators, or other agents during the learning process.
Based on beneficial relationships inherent in the data, \textbf{\textit{structurization with explicit graph-based modeling}} emerges as a promising way to address these challenges, \textit{transforming raw and complex inputs into concise and well-organized forms} that are more amenable to agent comprehension and processing. Such organized information is helpful for AI agents to explore complex tasks and make well-informed decisions.

Graphs have demonstrated extensive applicability in diverse domains~\cite{guo2020survey,ektefaie2023multimodal,chen2024survey} and have proved to be a powerful paradigm for managing data and organizing information with valuable relationships. Based on built graphs, graph learning then demonstrates success in using structured information~\cite{wu2020comprehensive,wu2023graph}. 
Specifically, forming a graph by representing and modeling entities as nodes and their explicit or implicit interconnections as edges provides an effective way for \textit{data organization}. 
A suitable graph is central to data organization for agents. The flexibility of graph construction allows for customized structures based on specific environments, tasks, operators, and applications~\cite{gallici2023transfqmix,zhuge2024gptswarm,liu2024toolnet}, or the usage of existing external knowledge graphs~\cite{yasunaga2021qa,zhao2024kg}. This versatility enables graphs to be widely integrated into various types of agents and their diverse functions.
Under the foundation of well-formed graphs, graph learning techniques can further provide a powerful framework with \textit{knowledge extraction} on graphs for capturing elaborate relationships and meaningful information for agents.
This makes the graph technique an ideal methodology for augmenting the capabilities of AI agents in complex scenarios. Consequently, the novel intersection of graphs and AI agents holds a significant promise for enhancing agent capacity to process and utilize structured information, thereby fostering more effective agent functionalities, such as planning, execution, memory, and multi-agent coordination.

\textbf{Taxonomy}. This survey systematically investigates the role of \textbf{graphs in the organization of information, operators, and multi-models for AI agents}, covering the spectrum from RL-based to LLM-based agent paradigms. Recognizing the increasing fusion of RL techniques with LLM foundation models, our analysis does not explicitly differentiate the role of graph learning between these agent architectures.
Instead, as illustrated in Figure~\ref{fig:overall}, our discussion is structured around \textbf{core agentic functionalities} empowered with graphs. We primarily focus on the potential of graph learning to bolster \textit{agent planning}, \textit{agent execution}, \textit{agent memory}, and \textit{multi-agent coordination}. Furthermore, this survey also explores the \textit{reciprocal relationship}, detailing how AI agents can, in turn, empower and refine graph learning processes. Finally, based on the reviews, we outline promising \textit{applications} and identify key \textit{future research opportunities} within this dynamic interdisciplinary field.
By reviewing comprehensive insights in this evolving field, we aim to catalyze new advancements in AI agents capable of leveraging structured knowledge to tackle increasingly sophisticated challenges.

Within the contexts in this article, existing surveys have primarily explored its utility in reinforcement learning~\cite{munikoti2023challenges,liu2024graph}.
More recently, amidst the significant advancements in LLMs, graph learning has also been recognized as an effective technique to improve LLM capabilities. Several reviews have covered this area~\cite{li2024survey,jin2024large}.
However, despite these contributions, a systematic review specifically elucidating how graphs can benefit various functionalities of an AI agent is notably absent. To the best of our knowledge, this survey presents the \textit{first systematic exploration} dedicated to the intersection of graph techniques and the multifaceted operations of AI agents. Based on this comprehensive review, our aim is to provide helpful insights that can inspire the development of next-generation AI agents.
The main contributions in the paper are listed as follows:
\begin{itemize}
    \item In this paper, we provide the first comprehensive review of the powerful and promising intersection between graph techniques and AI agents\footnote{Related resources are collected in the \myherf{\href{https://github.com/YuanchenBei/Awesome-Graphs-Meet-Agents}{\textbf{Github link}}}.}.
    \item We introduce a novel taxonomy that provides a structured perspective on the role graphs play across different agentic functionalities: planning, execution, memory, and coordination with other agents. We also discuss how agents, in turn, can advance graph learning.
    \item Based on the survey, we further discuss meaningful applications, key challenges, and future opportunities of graph-empowered AI agents.
\end{itemize}

The following sections are organized as follows. We give key preliminaries related to this survey in Section~\ref{sec:pre}. Then, a detailed related work survey based on our taxonomy is introduced in Section~\ref{sec:plan} to Section~\ref{sec:agent4graph}. In addition, promising applications and future opportunities for the intersection of graph techniques and AI agents are discussed in Section~\ref{sec:app} and Section~\ref{sec:future}. Finally, the key summary of this paper is presented in Section~\ref{sec:conclusion}.

\section{Preliminaries}\label{sec:pre}
\subsection{AI Agents}
An AI agent is an intelligent model capable of perceiving its environment and making autonomous decisions to achieve specific goals~\cite{wooldridge1995intelligent}. 
In traditional AI agents, a learnable model is mainly optimized by reinforcement learning. Through a carefully designed reward function, it is guided to make suitable decisions for a specific task as it interacts with the environment via a potential action set. Hence, we can also refer to this type of AI agent as an \textbf{RL agent}.

In recent years, with the significant advancements in generative AI centered on large language models, there has been a growing focus on agents based on LLMs.
We can refer to this newly emerging paradigm of agents as the \textbf{LLM agent}.
In addition to the ability to perceive the environment, understand tasks, and plan solutions for tasks, such as the previous RL agents, LLM agents also have memory mechanisms and tool usage abilities. They can store and manage memories and invoke external tools/functions to perform tasks.

Note that since reinforcement learning has already shown its powerful ability to enhance the large language model~\cite{cao2024survey,wang2024reinforcement,guo2025deepseek}, they are gradually merging into an inseparable whole. Therefore, in this survey, we will review the functionality of graph learning in AI agents from the early days of AI agents onward, without explicitly distinguishing between these two types of technical backbones.

\subsection{Reinforcement Learning}
Reinforcement Learning (RL) sits at the intersection of machine learning, control theory, and cognitive science. Unlike supervised learning, which relies mainly on labeled large datasets, RL lets an agent learn by acting, observing the consequences, and using scalar rewards or penalties to refine its behavior flexibly with or without human labels. The paradigm focuses on sequential decision making, that is, optimizing a chain of actions for long-term return~\cite{arulkumaran2017deep,moerland2023model}.

\textbf{Why RL is a suitable learning paradigm for AI agents?} Firstly, reinforcement learning is inherently built on an ''interaction-feedback'' loop, reflecting an agent's continuous ''perceive-act'' cycle. In this way, it can learn from experiences without relying heavily on labeled datasets. Secondly, RL pairs naturally with fast simulators, allowing agents to accumulate vast experience through self-play or exploration before deployment. Furthermore, RL encodes task objectives directly through reward design, letting the agent focus solely on maximizing return. 
These enable RL agents to flexibly design customized learning strategies according to the characteristics of specific tasks and environments.

\subsection{Large Language Models}

In recent years, based on extensive world knowledge during large-scale pretraining, large language models (LLMs) have demonstrated great knowledge expression ability in natural language understanding and generation tasks~\cite{chang2024survey}. 
Although RL can flexibly shape agents for different scenarios through reward-based learning, RL agents need to be trained individually for each specific task. LLM can serve as an agent's foundation model for various tasks without costly re-training, supported by the pre-trained foundational knowledge.

\textbf{Why LLMs are suitable foundation models for AI agents?} Firstly, the outstanding natural language understanding ability of LLMs allows them to accurately interpret user instructions and needs, forming the basis for effective agent-user interaction. Secondly, their powerful language generation enables them to create action strategies, call external tools/functions, and manage historical memory tailored to different task scenarios with textual descriptions and operations~\cite{guo2024large,zhang2024survey}. Furthermore, LLMs can continuously learn and update their knowledge through retrieval-augmented generation or post-training~\cite{fan2024survey,kumar2025llm}, adapting to diverse and changing tasks. This adaptability and capacity for evolution equip them to better serve users in various applications.

\tikzstyle{my-box}=[
    rectangle,
    draw=hidden-draw,
    rounded corners,
    align=center,
    text opacity=1,
    minimum height=1.5em,
    minimum width=5em,
    inner sep=2pt,
    fill opacity=.8,
    line width=0.8pt,
]
\tikzstyle{leaf-head}=[my-box, minimum height=1.5em,
    draw=gray!80, 
    fill=gray!15,  
    text=black, font=\normalsize,
    inner xsep=2pt,
    inner ysep=4pt,
    line width=0.8pt,
]

\tikzstyle{leaf-datasets}=[my-box, minimum height=1.5em,
    draw=myred!80, 
    fill=myred!15,  
    text=black, font=\normalsize,
    inner xsep=2pt,
    inner ysep=4pt,
    line width=0.8pt,
]

\tikzstyle{leaf-methods}=[my-box, minimum height=1.5em,
    draw=myblue!80, 
    fill=myblue!15 ,  
    text=black, font=\normalsize,
    inner xsep=2pt,
    inner ysep=4pt,
    line width=0.8pt,
]
\tikzstyle{leaf-metrics}=[my-box, minimum height=1.5em,
    draw=myorange!80, 
    fill=myorange!15,  
    text=black, font=\normalsize,
    inner xsep=2pt,
    inner ysep=4pt,
    line width=0.8pt,
]
\tikzstyle{leaf-app}=[my-box, minimum height=1.5em,
    draw=mygreen!80, 
    fill=mygreen!15,  
    text=black, font=\normalsize,
    inner xsep=2pt,
    inner ysep=4pt,
    line width=0.8pt,
]
\tikzstyle{leaf-chall}=[my-box, minimum height=1.5em,
    draw=mygrey!80, 
    fill=mygrey!15,  
    text=black, font=\normalsize,
    inner xsep=2pt,
    inner ysep=4pt,
    line width=0.8pt,
]

\tikzstyle{modelnode-datasets}=[my-box, minimum height=1.5em,
    draw=myred!80, 
    fill=white,  
    text=black, font=\normalsize,
    inner xsep=2pt,
    inner ysep=4pt,
    line width=0.8pt,
]

\tikzstyle{modelnode-methods}=[my-box, minimum height=1.5em,
    draw=myblue!100, 
    fill=white,  
    text=black, font=\normalsize,
    inner xsep=2pt,
    inner ysep=4pt,
    line width=0.8pt,
]
\tikzstyle{modelnode-metrics}=[my-box, minimum height=1.5em,
    draw=myorange!100, 
    fill=white,  
    text=black, font=\normalsize,
    inner xsep=2pt,
    inner ysep=4pt,
    line width=0.8pt,
]
\tikzstyle{modelnode-app}=[my-box, minimum height=1.5em,
    draw=mygreen!100, 
    fill=white,  
    text=black, font=\normalsize,
    inner xsep=2pt,
    inner ysep=4pt,
    line width=0.8pt,
]
\tikzstyle{modelnode-chall}=[my-box, minimum height=1.5em,
    draw=mygrey!100, 
    fill=white,  
    text=black, font=\normalsize,
    inner xsep=2pt,
    inner ysep=4pt,
    line width=0.8pt,
]
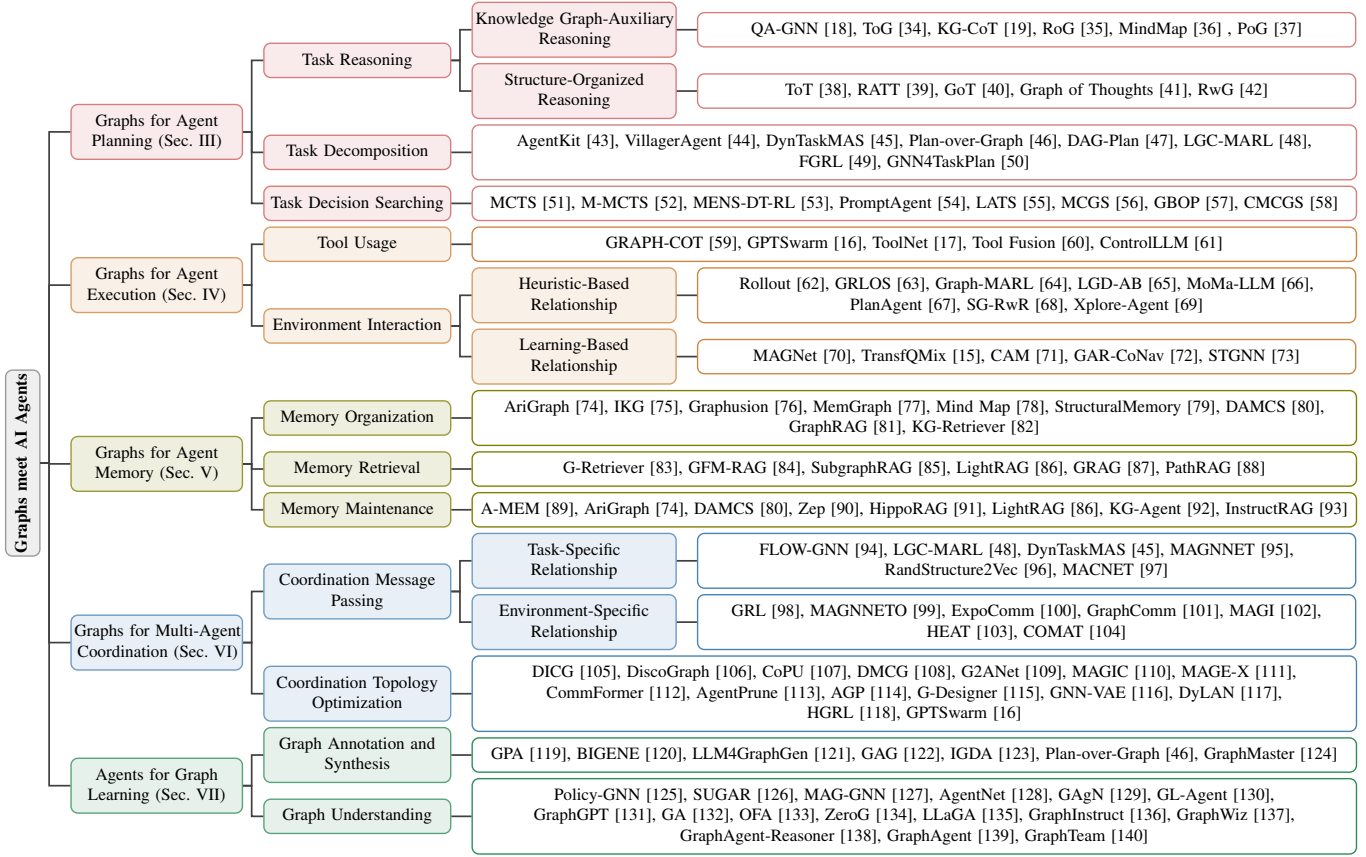
\begin{figure*}[!th]
    \centering
    \resizebox{1\textwidth}{!}{
        \begin{forest}
            forked edges,
            for tree={
                grow=east,
                reversed=true,
                anchor=base west,
                parent anchor=east,
                child anchor=west,
                base=left,
                font=\larger ,
                rectangle,
                draw=hidden-draw,
                rounded corners,
                align=center,
                minimum width=1em,
                edge+={darkgray, line width=1pt},
                s sep=3pt,
                inner xsep=0pt,
                inner ysep=3pt,
                line width=0.8pt,
                ver/.style={rotate=90, child anchor=north, parent anchor=south, anchor=center},
            }, 
            [
                \textbf{Graphs meet AI Agents}, leaf-head, ver
                [
                    Graphs for Agent \\ Planning (Sec.~\ref{sec:plan}), leaf-datasets, text width=9.6em
                    [ 
                        Task Reasoning, leaf-datasets, text width=10.5em
                        [
                            Knowledge Graph-Auxiliary \\Reasoning, leaf-datasets, text width=11.5em
                            [QA-GNN~\cite{yasunaga2021qa}{,} ToG~\cite{sun2024thinkongraph}{,} KG-CoT~\cite{zhao2024kg}{,} RoG~\cite{luo2024reasoning}{,} MindMap~\cite{wen2024mindmap} {,} PoG~\cite{tan2025paths},  modelnode-datasets, text width=38em]
                        ]
                        [
                            Structure-Organized \\Reasoning, leaf-datasets, text width=11.5em
                            [ToT~\cite{yao2023tree}{,} RATT~\cite{zhang2025ratt}{,} GoT~\cite{yao2024got}{,} Graph of Thoughts~\cite{besta2024graph}{,} RwG~\cite{han2025reasoning}, modelnode-datasets, text width=38em]
                        ]
                    ]
                    [
                        Task Decomposition, leaf-datasets, text width=10.5em
                            [AgentKit~\cite{wu2024agentkit}{,} VillagerAgent~\cite{dong2024villageragent}{,} DynTaskMAS~\cite{yu2025dyntaskmas}{,} Plan-over-Graph~\cite{zhang2025plan}{,} DAG-Plan~\cite{gao2024dag}{,} LGC-MARL~\cite{jia2025enhancing}{,}\\ FGRL~\cite{marzi2024feudal}{,} GNN4TaskPlan~\cite{wu2024can}, modelnode-datasets, text width=51.2em]
                    ]
                    [ 
                        Task Decision Searching, leaf-datasets, text width=10.5em
                        [MCTS~\cite{guo2014mcts}{,} M-MCTS~\cite{xiao2018memory}{,} MENS-DT-RL~\cite{costa2024evolving}{,} PromptAgent~\cite{wang2024promptagent}{,}  LATS~\cite{zhou2024language}{,} 
                        MCGS~\cite{czech2021improving}{,} GBOP~\cite{leurent2020monte}{,}  CMCGS~\cite{kujanpaa2024continuous}, modelnode-datasets, text width=51.2em]
                    ]
                ]
                [
                    Graphs for Agent \\Execution  (Sec.~\ref{sec:exec}), leaf-metrics,text width=9.6em
                    [ 
                        Tool Usage, leaf-metrics, text width=10.5em
                            [GRAPH-COT~\cite{jin2024graph}{,} GPTSwarm~\cite{zhuge2024gptswarm}{,} ToolNet~\cite{liu2024toolnet}{,} Tool Fusion~\cite{lumer2025graph}{,} ControlLLM~\cite{liu2024controlllm}, modelnode-metrics, text width=51.2em]
                    ]
                    [ 
                        Environment Interaction, leaf-metrics, text width=10.5em
                        [
                            Heuristic-Based \\ Relationship, leaf-metrics, text width=11.5em
                            [Rollout~\cite{garces2023multiagent}{,} GRLOS~\cite{johnn2024graph}{,} Graph-MARL~\cite{weil2024towards}{,} LGD-AB~\cite{huh2025multi}{,} MoMa-LLM~\cite{honerkamp2024language}{,}\\ PlanAgent~\cite{zheng2024planagent}{,} SG-RwR~\cite{chen2025schema}{,} Xplore-Agent~\cite{sun2025gui}, modelnode-metrics, text width=38em]
                        ]
                        [
                            Learning-Based\\Relationship, leaf-metrics, text width=11.5em
                            [
                            MAGNet~\cite{malysheva2018deep}{,} TransfQMix~\cite{gallici2023transfqmix}{,} CAM~\cite{yu2023learning}{,} GAR-CoNav~\cite{ou2024reinforcement}{,} STGNN~\cite{dik2024graph}, modelnode-metrics, text width=38em]
                        ]
                    ]
                ]
                [
                    Graphs for Agent\\ Memory (Sec.~\ref{sec:mem}), leaf-chall,text width=9.6em
                    [ 
                        Memory Organization, leaf-chall, text width=10.5em
                            [AriGraph~\cite{anokhin2024arigraph}{,}
                            IKG~\cite{zheng2021towards}{,} Graphusion~\cite{yang2024graphusion}{,} MemGraph~\cite{xiong2025enhancing}{,} Mind Map~\cite{wu2025agentic}{,} StructuralMemory~\cite{zeng2024structural}{,} 
                            DAMCS~\cite{yang2025llm}{,}\\
                            GraphRAG~\cite{edge2024local}{,} 
                            KG-Retriever~\cite{chen2024kg}, modelnode-chall, text width=51.2em]
                    ]
                    [ 
                        Memory Retrieval, leaf-chall, text width=10.5em
                            [G-Retriever~\cite{he2024g}{,} GFM-RAG~\cite{luo2025gfm}{,} SubgraphRAG~\cite{li2024subgraphrag}{,} LightRAG~\cite{guo2024lightrag}{,} GRAG~\cite{hu2024grag}{,} PathRAG~\cite{chen2025pathrag}, modelnode-chall, text width=51.2em]
                    ]
                    [
                        Memory Maintenance, leaf-chall, text width=10.5em
                            [A-MEM~\cite{xu2025mem}{,} AriGraph~\cite{anokhin2024arigraph}{,} DAMCS~\cite{yang2025llm}{,} Zep~\cite{rasmussen2025zep}{,}  HippoRAG~\cite{jimenez2024hipporag}{,}  LightRAG~\cite{guo2024lightrag}{,} KG-Agent~\cite{jiang2024kg}{,} InstructRAG~\cite{wang2025instructrag}, modelnode-chall, text width=51.2em]
                    ]
                ]
                [
                    Graphs for Multi-Agent \\Coordination (Sec.~\ref{sec:mas}), leaf-methods,text width=9.6em
                        [   Coordination Message \\Passing, leaf-methods, text width=10.5em
                            [
                                Task-Specific\\ Relationship, leaf-methods, text width=11.5em
                                [FLOW-GNN~\cite{zhang2025gnns}{,} LGC-MARL~\cite{jia2025enhancing}{,} DynTaskMAS~\cite{yu2025dyntaskmas}{,} MAGNNET~\cite{ratnabala2025magnnet}{,}\\ RandStructure2Vec~\cite{kang2022learning}{,} MACNET~\cite{qian2025scaling}, modelnode-methods, text width=38em]
                            ]
                            [
                                Environment-Specific\\Relationship, leaf-methods, text width=11.5em
                                [GRL~\cite{liu2022graph}{,} MAGNNETO~\cite{bernardez2023magnneto}{,} ExpoComm~\cite{li2025exponential}{,} GraphComm~\cite{shen2021graphcomm}{,} MAGI~\cite{ding2023robust}{,}\\ HEAT~\cite{mo2022multi}{,} COMAT~\cite{cai2024transformer}, modelnode-methods, text width=38em]
                            ]
                        ]
                        [
                            Coordination Topology\\Optimization, leaf-methods, text width=10.5em
                            [DICG~\cite{li2021deep}{,}  DiscoGraph~\cite{li2021learning}{,} CoPU~\cite{li2022online}{,} DMCG~\cite{gupta2025deep}{,} G2ANet~\cite{liu2020multi}{,} MAGIC~\cite{niu2021multi}{,} MAGE-X~\cite{yang2023learning}{,}\\ CommFormer~\cite{hu2024learning}{,} AgentPrune~\cite{zhang2025cut}{,} AGP~\cite{li2025adaptive}{,} G-Designer~\cite{zhang2024gdesigner}{,} GNN-VAE~\cite{meng2025reliable}{,}
                            DyLAN~\cite{liu2024dylan}{,}\\ HGRL~\cite{chen2024adaptive}{,} GPTSwarm~\cite{zhuge2024gptswarm}, modelnode-methods, text width=51.2em]
                        ]
                ] 
                [
                    Agents for Graph \\Learning (Sec.~\ref{sec:agent4graph}), leaf-app,text width=9.6em
                    [ 
                        Graph Annotation and \\Synthesis, leaf-app, text width=10.5em
                            [GPA~\cite{hu2020graph}{,} BIGENE~\cite{zhang2022batch}{,} LLM4GraphGen~\cite{yao2024exploring}{,} GAG~\cite{ji2024gag}{,} IGDA~\cite{havrilla2025igda}{,} Plan-over-Graph~\cite{zhang2025plan}{,} GraphMaster~\cite{du2025graphmaster}, modelnode-app, text width=51.2em]
                    ]
                    [ 
                        Graph Understanding, leaf-app, text width=10.5em
                            [Policy-GNN~\cite{lai2020policy}{,} SUGAR~\cite{sun2021sugar}{,} MAG-GNN~\cite{kong2023mag}{,} AgentNet~\cite{martinkus2023agentbased}{,} GAgN~\cite{liu2025graph}{,} GL-Agent~\cite{wei2023versatile}{,}\\ GraphGPT~\cite{tang2024graphgpt}{,} GA~\cite{wang2023graph}{,} OFA~\cite{liu2024one}{,} ZeroG~\cite{li2024zerog}{,} LLaGA~\cite{chen2024llaga}{,} GraphInstruct~\cite{luo2024graphinstruct}{,} GraphWiz~\cite{chen2024graphwiz}{,}\\ GraphAgent-Reasoner~\cite{hu2024scalable}{,} GraphAgent~\cite{yang2024graphagent}{,} GraphTeam~\cite{li2024graphteam}, modelnode-app, text width=51.2em]
                    ]
                ]
            ]
        \end{forest}
    }
    \caption{A taxonomy overview of this survey with representative methods on graph-empowered functionalities of AI agents and agent-facilitated graph learning.}
    \label{fig:taxonomy}
\end{figure*}

\subsection{Graphs and Graph Learning}
In the real world, explicit or implicit structural relationships are often available within the data. 
Graph techniques aim to build an effective graph model to organize the data with nodes (representing entities) and edges (representing relationships), thereby modeling or learning complex structured knowledge with an organized graph. They are widely used in fields such as social network analysis, bioinformatics, and recommendation systems~\cite{sankar2021graph,yi2022graph,chen2024macro}. 
The graph learning process can generally be divided into two key parts: data organization and knowledge extraction. The two parts can also answer \textbf{why graphs can play a powerful role in AI agents?}

\textbf{Data Organization (Graph)}.
A good graph structure is crucial for extracting effective information. 
Organizing data into a suitable graph is the foundation and an important part of information and operator structurization. 
Traditional AI agents mainly deal with unstructured data and need extensive environmental trials to gain task-relevant knowledge from complex task data. However, much of the information and operators that an agent encounters can be better organized via graphs. For instance, task planning can be facilitated using sub-task dependency graphs, memory can be organized through graphs, and communication patterns among multiple agents can also be represented as graph topologies.

\textbf{Knowledge Extraction (Graph Learning)}.
Once well-organized graphs are obtained, graph models, such as graph neural networks (GNNs), can extract task-required knowledge by leveraging and aggregating the information at each node and its neighborhood in the graph.
Compared to learning directly from unstructured data, defined nodes can share information with their neighbors when learning from structured knowledge, enabling them to make decisions not solely based on their local information.
In this way, it can more flexibly support an agent's complex operators, such as task dependency modeling, efficient decision search, effective memory management and retrieval, and multi-agent communication optimization.

\subsection{Taxonomy Overview}
In this paper, we present a novel taxonomy on how graph techniques and AI agents can enhance each other, as shown in Figure~\ref{fig:taxonomy}. 
Specifically, in Sections \ref{sec:plan}-\ref{sec:mas}, we introduce how graph learning can support core functionalities of agents, including planning (Section~\ref{sec:plan}), execution (Section~\ref{sec:exec}), memory (Section~\ref{sec:mem}), and multi-agent coordination (Section~\ref{sec:mas}). In this way, dividing the synergy points of graphs for agent functionalities not only matches the natural modularization in agent system design but also highlights the unique opportunities for graph-centric techniques in each functionality.
In addition, in Section~\ref{sec:agent4graph}, we also discuss how the agent paradigm can, in return, be used to strengthen the graph learning process. By explicitly considering the reverse direction, how agent-based paradigms can benefit graph learning, we emphasize bidirectional innovation and encourage a holistic view where graphs and agents co-evolve and deeply integrate, inspiring new methodologies that go beyond unidirectional adoption. 
Based on this well-organized taxonomy, we can then discuss applications and future opportunities in Sections~\ref{sec:app} and~\ref{sec:future}.

\section{Graphs for Agent Planning}\label{sec:plan}
\begin{figure*}[t]
\centering
\includegraphics[width=\linewidth]{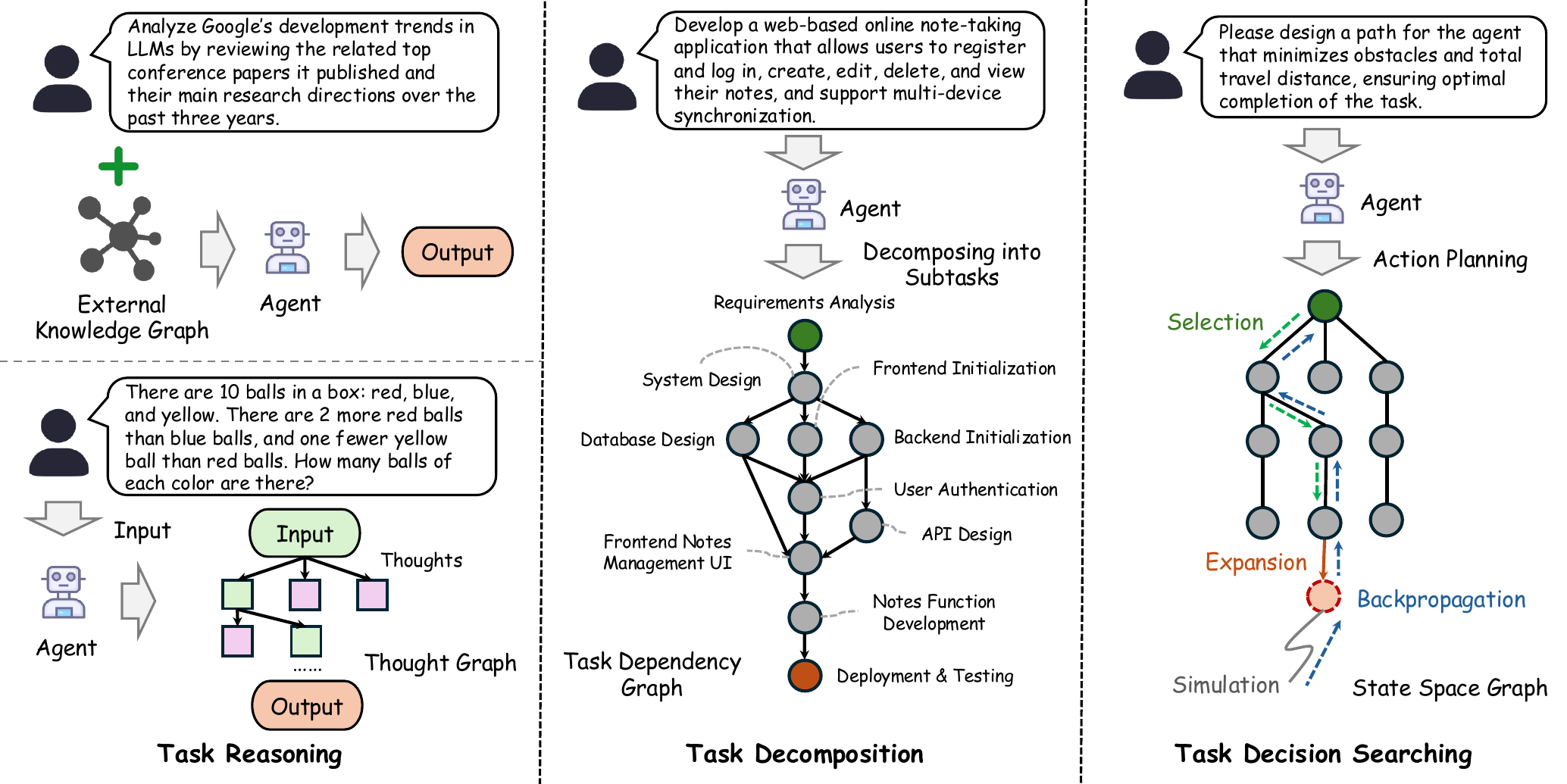}
%\vspace{-0.8em}
\caption{An illustration of graphs for agent planning.}
\label{fig:plan}
\end{figure*}

Planning is one of the foundational and crucial capabilities for an AI Agent. 
To achieve a specific goal or task, planning refers to the process in which an AI agent understands the task and devises a series of rational and ordered action plans after taking into account various available factors~\cite{huang2024understanding}.
For instance, in a game-playing scenario, the AI Agent needs to plan how to select appropriate equipment and determine the optimal route, etc., to complete the mission of passing through the challenges.
As illustrated in Figure~\ref{fig:plan}, graphs can play a role in various aspects of agent planning, mainly including effectively organizing the task reasoning form, arranging the task decomposition procedure, and constructing an efficient task decision searching process.

\subsection{Task Reasoning}
From Chain-of-Thought (CoT)~\cite{wei2022chain} to Deepseek-R1~\cite{guo2025deepseek}, many studies have improved prompts and thinking to boost the task reasoning of LLM-based agents. These approaches use the agents' intermediate thinking for deep task reasoning. To better enhance this thinking and help agents extract key task knowledge effectively with graphs, many works have explored the knowledge graph auxiliary and graph-based organization for the task reasoning process.

\subsubsection{Knowledge Graph-Auxiliary Reasoning}
Knowledge graph-assisted reasoning primarily enhances the agent's reasoning for tasks by leveraging additional information on an auxiliary knowledge graph. 
Formally, a \textbf{knowledge graph (KG)} is a structured representation of knowledge. The nodes in a KG denote entities or concepts that the model can recognize or generate, while the edges signify the relationships between these entities or concepts~\cite{ji2021surveykg}. 
The agent can extract multi-hop subgraph information from the auxiliary knowledge graph (KG) based on the entities and the relationships between entities involved in the task. This enables the agent to gain a deeper understanding of the entities and develop a more comprehensive grasp of the target task.

In the early period, a representative work QA-GNN~\cite{yasunaga2021qa} improves question-answering systems by combining language models and auxiliary KGs by using a relevance scoring mechanism to evaluate the importance of the KG node for the target tasks. Through GNNs that enable mutual representation updates, the model can handle complex structured reasoning tasks.
Subsequently, recent work has begun to utilize insights related to QA-GNN, given the significant success of using prompts to activate the capabilities of LLMs. 
For example, ToG~\cite{sun2024thinkongraph} and KG-CoT~\cite{zhao2024kg} have begun to enhance the task reasoning process of the LLM agent with KG-based retrieval-augmented generation (RAG). These studies focus on extracting key entities from prompts and employing graph retrieval techniques to identify related multi-hop subgraphs or paths of corresponding entities within auxiliary KGs. This process uncovers higher-order relationships and knowledge of the target task.
Subsequently, various methods have begun to explore enhancement mechanisms for the deeper integration between auxiliary KGs and LLM agents.
For example, RoG~\cite{luo2024reasoning} introduces a novel planning-retrieval-reasoning framework, focusing on generating relation paths from KGs as plans to optimize the reasoning process of LLM agents.
MindMap~\cite{wen2024mindmap} builds a mind map that more deeply connects external KG knowledge with the LLM's implicit knowledge for reasoning.
In addition, PoG~\cite{tan2025paths} includes dynamic multistage path exploration and efficient path pruning to extract multi-hop reasoning paths and information from KGs.

\subsubsection{Structure-Organized Reasoning}
Compared with directly generating responses, the intermediate thinking and reasoning processes of LLM agents can generate a wealth of additional helpful information. Structure-enhanced reasoning can help LLM agents understand key task-related knowledge more efficiently by structuring these processes with trees or more complex graph forms~\cite{chen2025towards}.

Initially, works explored using the tree structure, a special type of graph, to structurally enhance the CoT reasoning process.
Representatively, ToT~\cite{yao2023tree} constructs a tree-structured representation of coherent thoughts, enabling agents to explore multiple reasoning paths simultaneously and make global decision-making through self-evaluation of the current state.
RATT~\cite{zhang2025ratt} further adjusts and optimizes the structure of the thought tree by planning and looking ahead at each point in the thought branch to explore and evaluate multiple potential reasoning steps, combining the fact-checking ability of retrieval-augmented generation (RAG).
To consider more complex relationships in the reasoning process, GoT~\cite{yao2024got} and Graph of Thoughts~\cite{besta2024graph} represent information units of the LLM agent's thought as nodes and dependencies between them as edges to advance the agent's reasoning capabilities, which provides more complex relationships than a pure tree structure.
Furthermore, compared to GoT and Graph-of-Thought, which improve reasoning by organizing LLM-generated intermediate results into graph structures, RwG~\cite{han2025reasoning} extracts information from the context to build graphs. Then it continuously improves and optimizes these graphs through multiple rounds of verification, better aligning them with specific tasks.

\subsection{Task Decomposition}
In AI agents, task planning requires decomposing user requests (in natural language for LLM agents) into specific sub-tasks. A complex task usually consists of several subtasks that are easier to handle~\cite{huang2024understanding,guan2024intelligent}. 
Proper task decomposition is important because reasonable sub-task decomposition can improve the accuracy and efficiency of the agent in performing tasks.
These sub-tasks often have dependencies, such as one task's output serving as another's input.
These sub-tasks and their dependencies form a \textbf{task dependency graph (TDG)}. 
TDG is the primary organized graph structure used in the task decomposition process.
Formally, the nodes of a TDG are usually the several sub-tasks decomposed from a task. The features of the nodes can be described in natural language or key feature vectors of subtask nodes. The edges between sub-task nodes are strictly established based on dependencies such as task execution order. To ensure that no dead loops occur during task execution, a TDG is typically a directed acyclic graph (DAG).

There are two main ways to construct TDG. 
One is for tasks with \textbf{implicit subtask dependencies}, leveraging LLMs' reasoning like CoT to split subtasks and mine dependencies, such as DAG-Plan~\cite{gao2024dag}, LGC-MARL~\cite{jia2025enhancing}, and VillagerAgent~\cite{dong2024villageragent}. Note that the previous structure-organized reasoning methods can also be directly applied here for implicit subtask dependency modeling by prompting LLMs to perform task decomposition during reasoning. 
The other is to directly build a graph using \textbf{prior explicit subtask relationships and rules}, such as DynTaskMAS~\cite{yu2025dyntaskmas} and Plan-over-Graph~\cite{zhang2025plan}. 

Based on the determined TDG, the aim of agent planning is to \textit{identify connected, acyclic paths or subgraphs} in the TDG and sequentially invoke appropriate subtasks to complete the task.
Existing methods adopt mainly three main types of techniques, including \textbf{LLMs, RL, and GNNs}, to plan optimal execution paths or subgraphs in TDG.
Representatively, based on the reasoning ability of LLMs, AgentKit~\cite{wu2024agentkit}, VillagerAgent~\cite{dong2024villageragent}, and
Plan-over-Graph~\cite{zhang2025plan} describes the TDG textually for the LLM agent to generate the execution plan. 
Furthermore, DAG-Plan~\cite{gao2024dag} and LGC-MARL~\cite{jia2025enhancing} conduct reinforcement learning by designing specific reward functions on the TDG to complete the path planning.
Moreover, FGRL~\cite{marzi2024feudal} and GNN4TaskPlan~\cite{wu2024can} optimize a suitable sub-task execution path in the task graph using GNNs, with the information aggregation between subtask nodes.

\subsection{Task Decision Searching}
Task decision searching is an important process in the planning stage of AI agents, involving the making of sequential decisions in complex environments to achieve specific goals~\cite{ladosz2022exploration}.
Search algorithms involve state transitions within the decision space, which naturally form a graph structure with transitions between states. These states and their transitions form a \textbf{state space graph (SSG)}.
Formally, each node in the SSG represents a state, with the node's properties being the state's information, such as its textual description or parameters. The transitions between states serve as the edges between nodes. Note that the edge relationships in the SSG are typically optimized by the search algorithm itself, rather than being predefined. It involves an agent's process of exploring different states and actions to dynamically generate and update the SSG within an environment, based on the current state and environment, to identify a sequence of actions that maximizes long-term rewards.

Typically, Monte Carlo Tree Search (MCTS) is a heuristic search algorithm for decision-making~\cite{guo2014mcts}. It represents the state space of the complex problem in a hierarchical tree structure. Each node represents a state and each edge represents an action. 
This tree structure enables the algorithm to explore and manage possible states and actions in an orderly manner, which enhances the search efficiency and avoids blind searching in the vast state space.
Many approaches have used MCTS to improve the task-handling capability of agents. For example, PromptAgent~\cite{wang2024promptagent} uses MCTS for automatic prompt optimization. LATS~\cite{zhou2024language} provides a foundation for the unified planning and understanding of LLM agents using MCTS.
Some recent works have also explored the enhancement of agents' search abilities in tree-structured SSGs. For instance, M-MCTS~\cite{xiao2018memory} enhances the MCTS SSGs by incorporating a memory mechanism to boost online search performance for models. MENS-DT-RL~\cite{costa2024evolving} uses evolutionary algorithms to search tree structures, thus improving the interpretability of decision trees.
Some methods have found that tree-structured SSGs still suffer from computational inefficiency and information sharing issues. They cannot share evaluation info across branches, causing a waste of computation and memory. To solve this, these methods expand the search structure from a tree to a directed acyclic graph, such as MCGS~\cite{czech2021improving} and GBOP~\cite{leurent2020monte}, enabling the sharing of information between branches and merging of nodes with the same state representation for a more efficient search. 
Furthermore, CMCGS~\cite{kujanpaa2024continuous} enhances the SSG to extend MCGS from a search in discrete state spaces to an efficient search in continuous spaces.

\section{Graphs for Agent Execution}\label{sec:exec}
\begin{figure}[t]
\centering
\includegraphics[width=\linewidth]{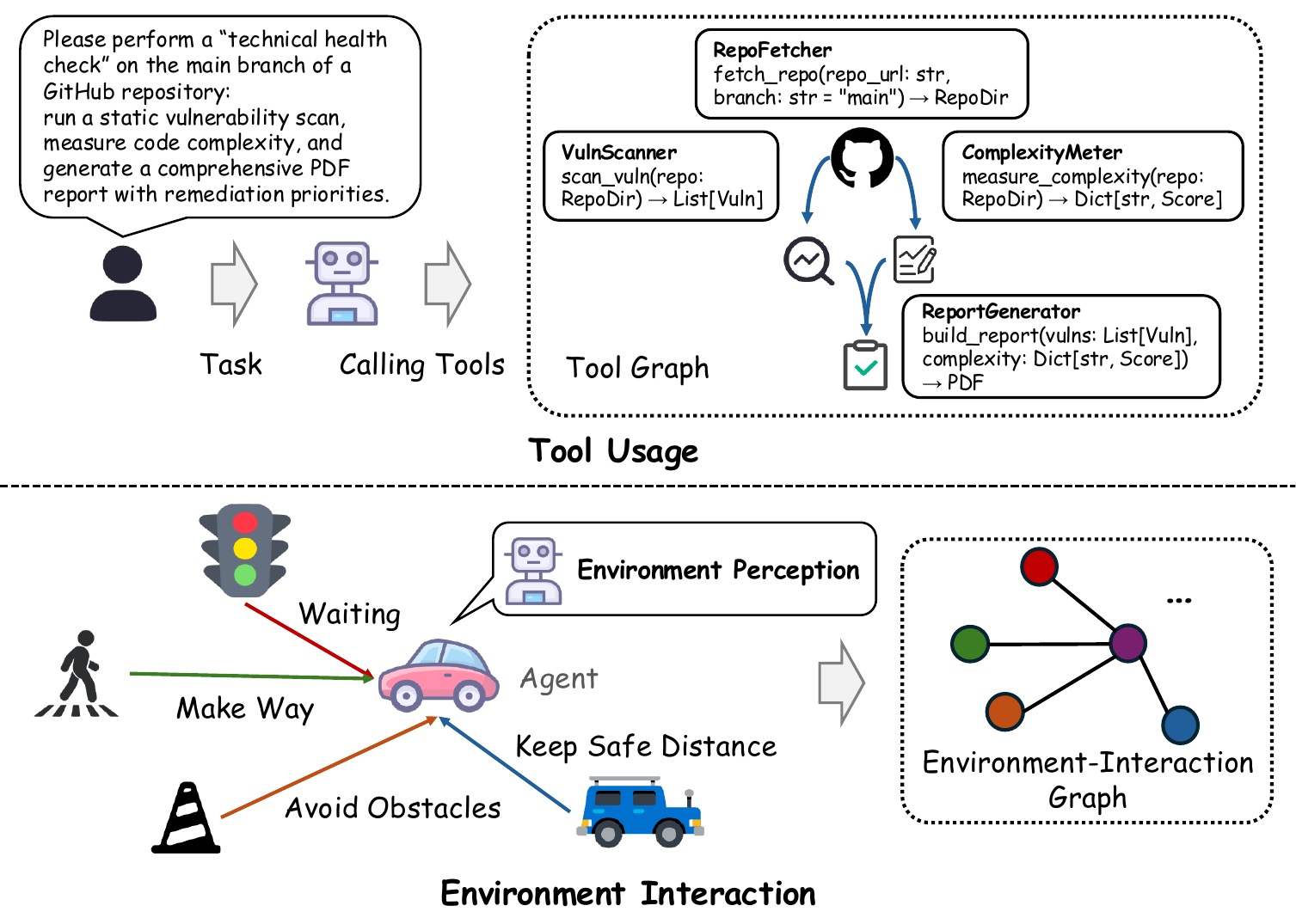}
%\vspace{-0.8em}
\caption{An illustration of graphs for agent execution.}
\label{fig:exec}
\end{figure}

After the planning of an agent, the execution phase is where the formulated plan is put into action. 
At this stage, two main modules can play an important role. (1) Tool usage: the agent needs to call upon appropriate external tools in combination with its own knowledge in order to complete the specified actions. (2) Environment interaction: the agent should interact with the neighboring environment to perceive the information it faces and conduct actions based on current circumstances. Graphs can help arrange the scheduling of these numerous tools and model the rich relationships between agents and the environments in which they reside. The examples for these processes are provided in Figure~\ref{fig:exec}.

\subsection{Tool Usage}
Agent tool usage refers to the process in which an agent invokes predefined tools, which can be APIs, software libraries, or various function calls, to perform specific actions. 
For example, GRAPH-COT~\cite{jin2024graph} strengthens the graph reasoning abilities of LLM agents by interactive tool calls with external graph-related knowledge during the reasoning process.
With the recent development of the Model Context Protocol (MCP), which is an open, unified, structured, and secure standard introduced by Anthropic\footnote{\url{https://github.com/modelcontextprotocol}} in 2024, the complex structured invocation of external data by agents has begun to attract widespread attention.
When an agent has a multitude of tools, there are often intricate dependencies and invocation relationships among them. 

In addition to helping agents reason graph-like data, graph organization and learning can also be used to construct a tool graph to clearly display the interconnections between these tools, enabling the agent to efficiently use and manage a vast number of tools.
To achieve this target, GPTSwarm~\cite{zhuge2024gptswarm} presents a flexible pipeline for graph-based tool usage, where an agent is abstracted as a DAG. In this framework, each node stands for a certain function, and the edges show the information flow between the nodes.
ToolNet~\cite{liu2024toolnet} and Tool Fusion~\cite{lumer2025graph} leverage the sparsity of tool usage patterns to organize tools into a directed graph, which means that when a particular tool is invoked, the subsequent tool to be called is usually restricted to a very limited set of options. This enables agents to handle a large number of tools without a significant increase in token consumption. In addition, ControlLLM~\cite{liu2024controlllm} constructs a tool graph containing resource nodes and tool nodes and employs parallel graph search algorithms to find suitable tool paths efficiently, avoiding the hallucination problem in tool usage.

\subsection{Environment Interaction}
As the environment serves as the foundation for an agent to perform tasks and optimize its behavior, the ability to perceive and interact with it is a key capability of the agent. 
Graphs serve as a general-purpose data structure for modeling relationships among agents and environmental entities, such as connectivity, spatial proximity, and interaction dependencies. 
A scene graph, a specialized type of graph, represents visual scenes by encoding objects as nodes and their spatial or semantic relationships (e.g., "on top of," "holding") as edges~\cite{chang2021comprehensive}. 
Approaches to modeling these relationships can be divided into two categories: heuristic-based methods, which rely on predefined rules or prior knowledge, and learning-based methods, which learn relationships through data-driven optimization.

\subsubsection{Heuristic-Based Relationship}
Heuristic-based relationships are manually defined based on prior knowledge of the environment and are widely adopted in agent systems.

RL-based agents need to interact with the environment extensively to optimize policies through specially-designed RL algorithms. Modeling the relationships among agents and environmental entities is therefore crucial for the effectiveness of these algorithms.
Examples include self-driving agents like Rollout~\cite{garces2023multiagent} in urban environments, embodied agents such as GRLOS~\cite{johnn2024graph} in warehouse-scheduling settings, routing agents like Graph-MARL~\cite{weil2024towards} in communication network environments, and auto-bidding agents such as LGD-AB~\cite{huh2025multi} in online advertising scenarios. 
Once these graphs are designed for specific environments, the corresponding relational structures can serve as structural priors to assist RL-based agents in interacting with the environment and achieving higher rewards. 

More recently, the advent of LLMs with strong reasoning and planning capabilities has led to growing interest in their integration into agent systems.
However, LLM-based agents often struggle to ground their abilities in situated environments due to the lack of environmental representations that LLMs can effectively process~\cite{huang2023grounded}.
Scene graphs~\cite{chang2021comprehensive} provide structured representations of the environment that complement the reasoning abilities of LLMs, thus enhancing the environmental understanding and interaction abilities of LLM-based agents.
This synergy has been explored in various settings, including embodied agents such as MoMa-LLM~\cite{honerkamp2024language} in indoor environments, self-driving agents such as PlanAgent~\cite{zheng2024planagent} in outdoor environments, and LLM agents like SG-RwR~\cite{chen2025schema} in simulation environments. 
Similarly, Xplore-Agent~\cite{sun2025gui} constructs a GUI Transition Graph to model the relationships between GUI screens, guiding LLMs through multiple downstream tasks.

\subsubsection{Learning-Based Relationship}
Since heuristic-based relationships rely heavily on manually designed ones and often struggle to generalize across environments, some works have started to suggest learning-based relationships.

These methods involve initializing relationships and then learning the relationships with learnable parameters or specific models, which is more flexible in different environments.
After determining the node content and initial node relationships, these methods dynamically learn the relationship weights or embeddings between nodes using learnable parameters or attention-based models like Transformer~\cite{vaswani2017attention} or Graph Attention Networks (GAT)~\cite{velivckovic2018graph}. This helps capture the relationship between the agent and environment nodes. 
For example, MAGNet~\cite{malysheva2018deep} and TransfQMix~\cite{gallici2023transfqmix} initialize graphs based on entities in the environment that the agents depend on or observe, then use self-attention mechanisms to dynamically adjust edge weights and determine neighborhood importance. 
CAM~\cite{yu2023learning} first constructs edge features based on the obstacle-rich environment and then uses a multi-layer perceptron to further learn the constructed edges.
GAR-CoNav~\cite{ou2024reinforcement} identifies node entities in the global environment and uses GAT to learn the association strength between nodes.
STGNN~\cite{dik2024graph} models the environment as a grid-based graph and incorporates learnable weights into Graph Convolutional Networks (GCN)~\cite{kipf2017semisupervised} to capture spatial and temporal dependencies within the data.

\section{Graphs for Agent Memory}\label{sec:mem}
\begin{figure*}[t]
\centering
\includegraphics[width=\linewidth]{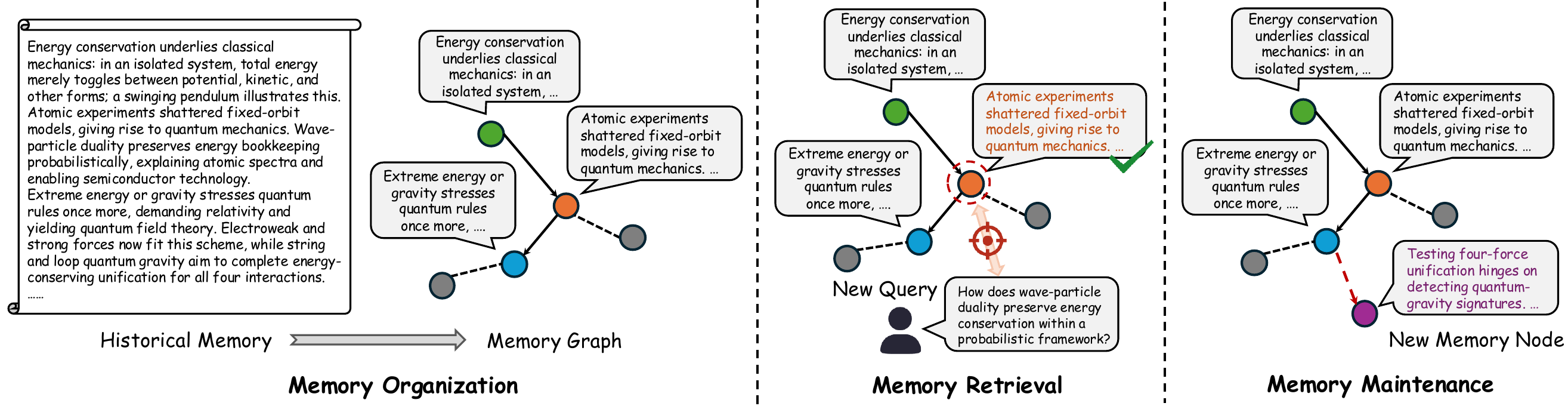}
%\vspace{-0.8em}
\caption{An illustration of graphs for agent memory.}
\label{fig:memory}
\end{figure*}

Memory is a crucial capability that allows agents to store and recall past experiences or related knowledge~\cite{luo2025large, zhang2025personaagent}. It allows agents to accumulate experience, thus facilitating more informed and appropriate actions.
When applied to agent memory, a graph-based memory organization can effectively uncover latent associations among the various information encountered by the agent. Based on a graph-based memory organization, equipped with effective memory retrieval and maintenance, the agent's ability is enhanced to recognize patterns and connections that might not be immediately apparent, thereby enriching its knowledge base and improving its execution in dynamic environments over an extended period. A toy example of graph-empowered memory is demonstrated in Figure~\ref{fig:memory}.

\subsection{Memory Organization}

Agents equipped with graph-structured memory can store knowledge and experiences as interconnected representations, inspired by modern cognitive models of memory that emphasize relational and contextual reasoning capabilities in humans~\cite{kumaran2016learning, griffiths2019doing}.
Recent LLM-based agents have explored a spectrum of memory representations, from unstructured text chunks to structured knowledge graphs. In particular, knowledge graphs (composed of semantic triples) and other structured forms (e.g., atomic facts and summarized notes) are increasingly used to organize an agent’s long-term memory.
For example, the AriGraph~\cite{anokhin2024arigraph} system represents an agent’s memory as a structured knowledge graph of triples that encodes both semantic facts and episodic events from the agent’s experience. 
Such graph-based memory organization enables the agent to recall not just isolated data points but the structured connections between them (e.g. cause–effect relations, temporal or relational links), which is essential for different scenarios like IKG~\cite{zheng2021towards} for industrial multimodal data, Graphusion~\cite{yang2024graphusion} for education, MemGraph~\cite{xiong2025enhancing} for patent analysis, and Mind Map~\cite{wu2025agentic} for web search. 
Moreover, StructuralMemory~\cite{zeng2024structural} evaluates four types of memory structures: chunks, knowledge triples, atomic facts, and summaries, in multiple scenarios. The study found that different memory structures offer distinct advantages tailored to specific tasks.

Integrating richer information, some works begin to introduce \textbf{hierarchical or hybrid graphs} with multilevel or multigranularity information to improve organization. For instance, a decentralized multi-agent framework, DAMCS~\cite{yang2025llm}, uses a hierarchical knowledge graph as memory, where multi-modal experiences are organized at different levels of abstraction. This hierarchy allows agents to record low-level details and high-level summaries and to share only the relevant portions with teammates, enhancing cooperative behavior.
Similarly, GraphRAG~\cite{edge2024local} constructs a hybrid knowledge graph that encodes both entity- and community-level nodes. This layered design enables the retrieval of tightly connected local facts while preserving broader semantic clusters, facilitating context-aware responses to questions requiring both fine-grained and holistic understanding.
KG-Retriever~\cite{chen2024kg} further advances this direction by indexing knowledge into a hierarchical graph composed of entity-level and document-level nodes. This structure facilitates follow-up retrieval at different semantic resolutions, enabling the agent to reason over specific entities or to aggregate information from entire documents depending on the complexity of the query.

\subsection{Memory Retrieval}
Based on the structured memory, it is critical for LLM agents to accurately and efficiently retrieve useful information from it as reliable guidance. 
Although traditional similarity-based retrievers can be directly applied to graph-based memory~\cite{robertson2009probabilistic}, to fully leverage the characteristics of graph information, recent works, mainly including graph-based retrieval-augmented generation (graph-based RAG), have paid attention to the retrievers specifically designed for graph-based knowledge bases and memories~\cite{peng2024graph,zhang2025survey}.

Among the process of the retrieval, some works focus on the \textbf{accuracy aspect} of retrieved knowledge subgraphs, which are the key factors that directly influence the decision made by agents. Representatively, G-Retriever~\cite{he2024g} and GFM-RAG~\cite{luo2025gfm} integrate the semantic similarity and the graph metrics for customized retriever design to rank and search the informative subgraphs, while SubgraphRAG~\cite{li2024subgraphrag} provides a lightweight perceptron with the triple-scoring mechanism for flexible subgraph retrieval based on the query relevance. Furthermore, LightRAG~\cite{guo2024lightrag} proposes a dual retriever system, including a local retriever for entity-level questions and a global retriever for complex queries requiring subgraph reasoning.

In addition to the accuracy, as the retrieval of memories demands extra processing time that affects the efficiency of agents, some studies also conduct the \textbf{efficiency-oriented} optimization for the retrieval with balancing effectiveness. For example, GRAG~\cite{hu2024grag} proposes a novel divide-and-conquer strategy to retrieve the optimal subgraph in linear time, along with two complementary views to help LLMs understand the graph context. In addition, PathRAG~\cite{chen2025pathrag} removes semantically overlapping paths with flow-based pruning, reducing the retrieval latency while maintaining the recall score.

\subsection{Memory Maintenance}

Compared with graph-based RAG, beyond organizing and retrieving structured knowledge, a pivotal and challenging aspect of graph-based agent memory is its dynamic evolution, the capacity to continuously update and refine memory representations and graph topologies in response to new experiences and interactions. 

Therefore, \textbf{dynamic graph}-based memory architectures have emerged in some studies as effective solutions for maintaining and evolving long-term memory in LLM agents~\cite{hatalis2023memory}. 
For example, A-MEM~\cite{xu2025mem} creates interconnected knowledge networks through dynamic indexing and linking, allowing continuous refinement of contextual representations as new memory is integrated. AriGraph~\cite{anokhin2024arigraph} consists of episodic and semantic memories. Episodic memory records observations received by the agent at specific steps and their temporal relationships. The agent continuously detects and removes outdated knowledge, expands semantic memory, and updates episodic memory to maintain the dynamic graph.

To better capture and store the long-term relationship of elements, some works have further constructed memories through \textbf{hierarchical dynamic graphs}.
Representatively, DAMCS~\cite{yang2025llm} proposes a goal-oriented hierarchical knowledge graph for long-term memory, with lower-level experience nodes and goal nodes for the agent's journey tracking. Higher-level long-term goal nodes are generated to provide an overview of the long-term progress.
Zep~\cite{rasmussen2025zep} is supported by Graphiti, a temporal-aware hierarchical knowledge graph engine.
It dynamically integrates unstructured conversational data, maintaining historical relationships. Memory management is achieved via a multi-layer knowledge graph with plot, semantic entity, and community subgraphs. For graph updating strategies, some works like HippoRAG~\cite{jimenez2024hipporag} and LightRAG~\cite{guo2024lightrag} include dynamically incremental graph updating. KG-Agent~\cite{jiang2024kg} introduces LLMs for updating the knowledge graph. Furthermore, InstructRAG~\cite{wang2025instructrag} adopts an RL agent with designed actions to maintain and update graphs.

\section{Graphs for Multi-Agent Coordination}\label{sec:mas}

\begin{figure}[t]
\centering
\includegraphics[width=\linewidth]{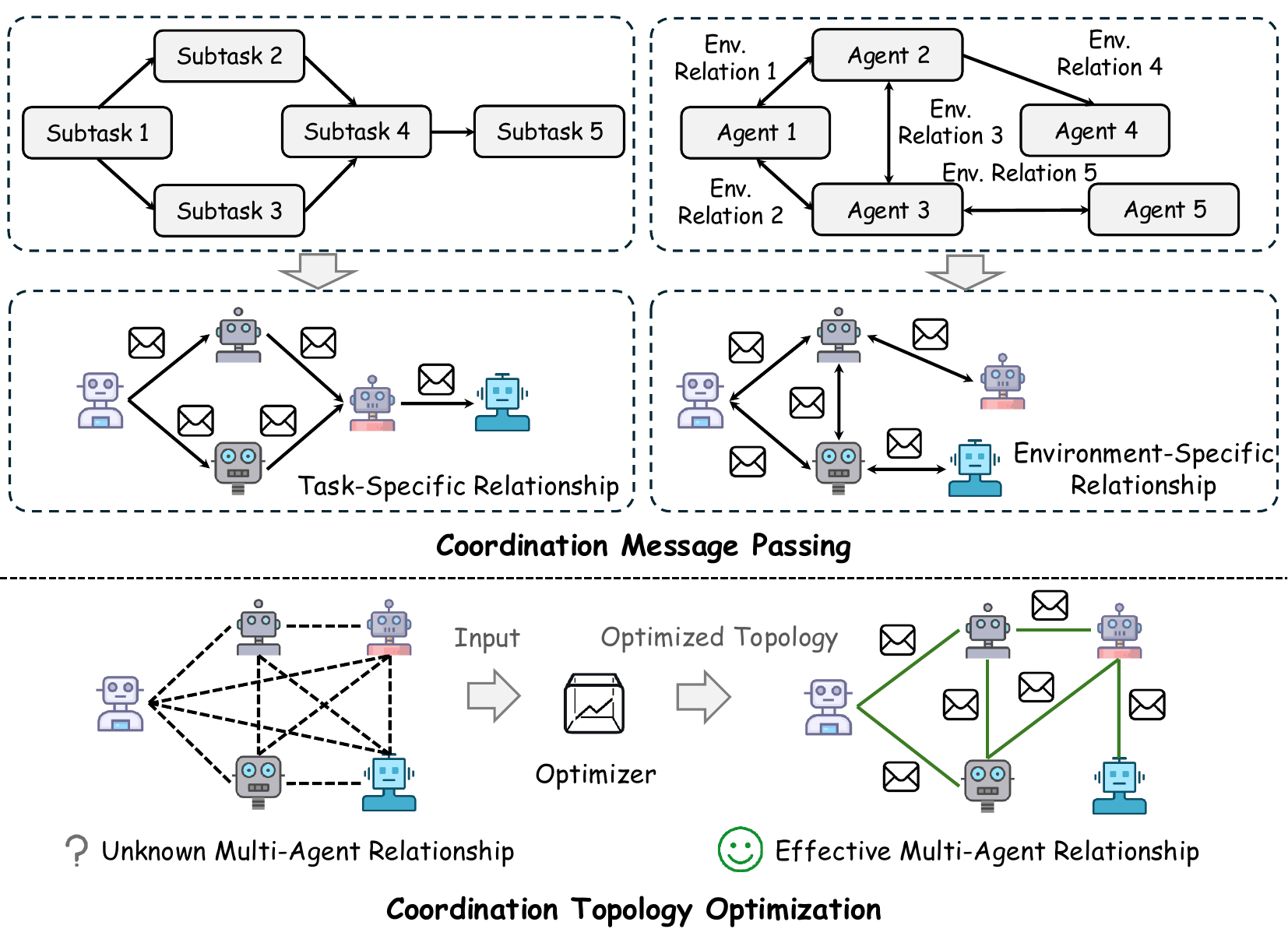}
%\vspace{-0.8em}
\caption{An illustration of graphs for multi-agent coordination.}
\label{fig:mas}
\end{figure}

A multi-agent system (MAS) is a complex system that integrates multiple AI agents for mutual integration, collaboration, and competition. Its goal is to accomplish complex tasks that are difficult for a single agent to complete through the interaction between various agents~\cite{amirkhani2022consensus,han2024llm}. In these systems, each agent has its own specialized advantages, such as domain-specific knowledge, reasoning ability, and the overall effect of coordination depends on the information exchange and decision consensus between the agents.
The core focus of coordinating multiple agents is relational modeling. Graph organization and learning can take advantage of its inherent ability to process relational data to excel in such modeling.
To make the review consistent, we define the graph organization of multi-agent coordination as the \textbf{agent coordination graph (ACG)}. Formally, in the ACG, the core nodes are agents. Node features represent the information of each agent. Edges are communication paths between agent nodes that are formed to pass messages between agents. For an MAS, graphs have mainly two roles. First, based on the predefined ACG (task-specific and environment-specific relationships), learn what information agents should communicate to maximize the MAS's performance. Second, optimize the ACG topology to improve communication effectiveness among multiple agents. The process is demonstrated in Figure~\ref{fig:mas}.

\subsection{Coordination Message Passing}
Heuristic relationships embedded in the tasks agents perform and in the environments with which they interact can be leveraged to construct the interaction topology of a multi-agent system. Building on this prior relational knowledge, recent work has begun to explore how to transmit and optimize messages among agents more effectively.

\subsubsection{Task-Specific Relationship}
The task-specific relationship consists mainly of two aspects: explicit task dependencies and task allocation to agents through implicit relationships.

\textbf{Task Dependency-Based Relationship}. 
In this type of relationship, the TDG for subtask dependency modeling is used as the ACG.
Within the agentic task execution workflow, multiple agents, respectively, take charge of a subtask. These agents will conduct message passing for efficient task execution based on the TDG's dependency relationship. For instance, based on the TDG, FLOW-GNN~\cite{zhang2025gnns} and LGC-MARL~\cite{jia2025enhancing} adopt GNNs and Graph-based multi-agent RL, respectively, to optimize the task workflow. DynTaskMAS~\cite{yu2025dyntaskmas} proposes a framework for dynamically managing multi-agent collaboration under TDG. Moreover, MACNET~\cite{qian2025scaling} uses the critic-actor task DAG for agent collaboration.

\textbf{Task Allocation-Based Relationship}. 
For multiple agents that need to collaborate to complete a complex task in an environment without explicit inter-subtask dependencies, some methods initiate the graph construction of a task allocation-based relationship for effective information extraction and exchange, such as RandStructure2Vec~\cite{kang2022learning} and MAGNNET~\cite{ratnabala2025magnnet}. Specifically, these methods introduce task-representing nodes in the ACG.
In this way, by connecting agent nodes with relevant task nodes and leveraging multi-layer graph message passing with GNNs, they capture higher-order agent-agent, agent-task, and task-task associations, thereby enhancing multi-agent task processing abilities.

\subsubsection{Environment-Specific Relationship}
The environment-specific relationships in multi-agent coordination focus on modeling the relationships among multiple agents and message passing under a particular agent environment. In this kind of relationship, each node in the ACG represents an agent, and the graph construction is based on the characteristics of the specified environment.

Similarly to the heuristic-based relationships discussed in Section~\ref{sec:exec}, some approaches rely on predefined rules to represent the relationships among agents. For example, GRL~\cite{liu2022graph} models vehicles as graph nodes and their interactions as edges for decision-making in interactive traffic scenarios. In MAGNNETO~\cite{bernardez2023magnneto}, agents exchange information with neighboring agents according to the underlying graph structure in computer networks. ExpoComm~\cite{li2025exponential} uses the exponential topology for efficient information dissemination among agents in large-scale multi-agent systems. 
In contrast, similar to the learning-based relationships discussed earlier, several methods dynamically learn the weights of the relationship among agents after initialization in specific environments. For instance, GraphComm~\cite{shen2021graphcomm} and MAGI~\cite{ding2023robust} explore multi-agent communication in game-playing scenarios, HEAT~\cite{mo2022multi}  focuses on urban driving scenarios, and COMAT~\cite{cai2024transformer} targets cooperation in heterogeneous multi-robot systems.

\subsection{Coordination Topology Optimization}
Due to the diverse task settings in the real world, there is always a lack of available prior relationships for designing the multi-agent coordination graph structure. Therefore, it is necessary to learn and optimize a communication topology for a given multi-agent scenario~\cite{liu2024dylan,zhang2024gdesigner}. Many existing works have explored this issue and can be categorized into three main classes for topology optimization, including edge importance measurement, graph autoencoder optimization, and reinforcement learning.

\textbf{Edge Importance Measurement}. 
In an overall pipeline, these methods first initialize a fully connected or rule-based topology between multiple agents. Then, on the one hand, many works select helpful edges via attention mechanisms, importance scores, or learnable parameterized edge weights, like DICG~\cite{li2021deep}, DiscoGraph~\cite{li2021learning}, CoPU~\cite{li2022online}, and DMCG~\cite{gupta2025deep}.
Building on this line, as parameterized connected graphs are computationally inefficient due to the high topology density, some methods convert them back to discrete graphs by threshold binarization (edges exist if weights exceed a threshold, otherwise not) or differentiable Gumbel-Softmax discretization~\cite{jang2017categorical}, such as G2ANet~\cite{liu2020multi}, MAGIC~\cite{niu2021multi}, MAGE-X~\cite{yang2023learning}, and CommFormer~\cite{hu2024learning}. On the other hand, some works directly utilize trainable graph masking for redundant and unnecessary communication node or edge pruning. For example, AgentPrune~\cite{zhang2025cut} identifies key multi-agent connections through low-rank-principled graph masking, and AGP~\cite{li2025adaptive} dynamically determines which agent nodes participate in specific tasks through a node masking head.

\textbf{Graph Auto-Encoder Optimization}. 
Graph auto-encoders (GAEs) encode node representations via an encoder layer and optimize specific tasks via a decoder layer. Many graph learning works use the GAE architecture for self-supervised link prediction between nodes~\cite{salha2019gravity,hou2022graphmae}. The core idea is to learn node representations with the encoder and to predict the links with a decoder. Based on this insight, GAE can be used for agent communication topology learning by predicting the high-confidence edges that exist between agents. Related works like G-Designer~\cite{zhang2024gdesigner} and GNN-VAE~\cite{meng2025reliable}, through graph variational autoencoders~\cite{kipf2016variational}, robustly predict whether edges exist between agent nodes to determine the optimal multi-agent coordination topology.
AGP~\cite{li2025adaptive} employs GAE for soft-pruning on sampled communication topologies and supplying supervision signals for the subsequent optimization within the framework.

\textbf{Reinforcement Learning}.
RL-based approaches that optimize multi-agent ACG topologies focus primarily on crafting an appropriate reward function that decides whether an edge between two agent nodes should exist or not.
For example, HGRL~\cite{chen2024adaptive} uses the link agent to learn which edges to add or remove for a node to maximize long-term RL rewards. GPTSwarm~\cite{zhuge2024gptswarm} parameterizes edge probabilities as a continuous and optimizable distribution, updating it based on task outcomes such as task response accuracy. The main challenge of these RL methods lies in balancing modeling effectiveness, overhead, and efficiency.

\section{Agents for Graph Learning}\label{sec:agent4graph}

\begin{figure}[t]
\centering
\includegraphics[width=\linewidth]{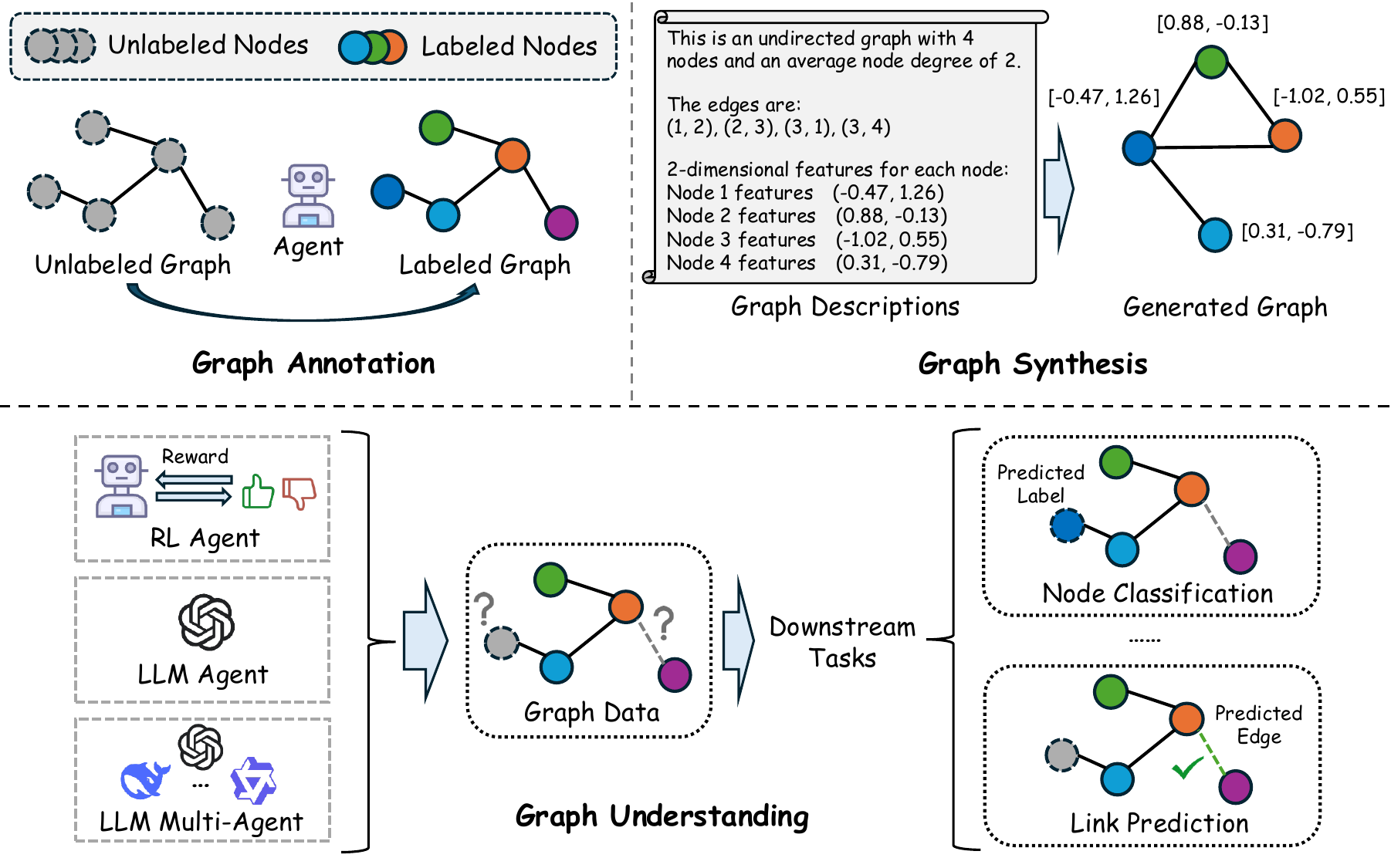}
\caption{An illustration of agents for graph learning.}
\label{fig:agent4graph}
\end{figure}

As shown in the previous sections, graph learning has contributed significantly to the capabilities of AI agents at various stages. 
However, the relationship between graphs and agents is not just unidirectional. Similarly, the paradigm of AI agents can also empower graph learning and graph-related tasks, opening up new avenues for automatic and effective graph processing. As illustrated in Figure~\ref{fig:agent4graph}, this synergy manifests itself primarily in two main aspects: (1) Graph annotation and synthesis, as well as (2) Graph understanding tasks.

\subsection{Graph Annotation and Synthesis}
Graph annotation and synthesis are crucial in graph learning due to the high cost of labeling and synthesizing high-quality graph data for graph model training in many real-world applications~\cite{zhu2022survey,huang2024cost}.

\textbf{Graph Annotation}. As graph annotation can be regarded as a decision-making process for label prediction and correction target, RL agents can be optimized for annotation through trial.
Representatively, GPA~\cite{hu2020graph} formulates the active learning problem on graphs as a Markov decision process and uses RL to learn the optimal query strategy.
GPA~\cite{hu2020graph} can be jointly trained on several source graphs with full labels to learn a transferable active learning policy, which can be directly generalized to unlabeled graphs.
BIGENE~\cite{zhang2022batch} models graph active learning as a multi-agent RL problem for node annotation. At each time step, each agent in BIGENE chooses an unlabeled node, forming a joint action. All agents share the same reward function and aim to collaborate to maximize the improvement of the classification model.

\textbf{Graph Synthesis}. With the abundant pre-trained knowledge of LLMs, many works have adopted the LLM agent paradigm to generate specific graphs for training or simulation. 
For instance, LLM4GraphGen~\cite{yao2024exploring} proposes a flexible pipeline where graph generation tasks are formulated as textual prompts, and LLM agents are required to output graphs in a specific format.
GAG~\cite{ji2024gag} adopts LLM agents to generate dynamic social graphs attributed to text. It uses agents to automatically simulate human-item interactions to produce graphs that mirror real-world networks.
IGDA~\cite{havrilla2025igda} is an iterative graph generation agent that predicts the initial graph by prompting the LLM to reason about the causal relationships and then updates the graph iteratively. In each round, edges with the highest uncertainty are prioritized for experimentation. After the experiment, the graph was updated based on the feedback.
Plan-over-Graph~\cite{zhang2025plan} first synthesizes an initial graph with defined rules. Then, an LLM agent is prompted to convert the graph into a description, with self-correction to ensure consistency between the generated graph and the original graph.
GraphMaster~\cite{du2025graphmaster} further proposes a multi-agent framework for graph data synthesis, which comprises four specialized LLM agents: the manager, perception, enhancement, and evaluation agents, that collaborate to optimize the synthesis process.

\subsection{Graph Understanding}
Traditional graph learning mainly aims to learn suitable representations and model parameters for graph understanding and graph modeling tasks, including node classification, link prediction, and graph classification.

\textbf{RL Agents}. Traditional graph modeling methods, such as GCN~\cite{kipf2017semisupervised} and GAT~\cite{velivckovic2018graph}, are based on fixed design logic. However, real-world graph structures are diverse, and fixed modeling logic cannot fully utilize the potential of graph learning. Therefore, methods like RL agents have started to design adaptive aggregation mechanisms for graph learning. For example, Policy-GNN~\cite{lai2020policy} uses RL to adaptively determine the number of aggregations for each node, addressing the issue that different nodes require varying aggregation iterations to adequately capture structural information. SUGAR~\cite{sun2021sugar} and MAG-GNN~\cite{kong2023mag} employ RL to design adaptive subgraph-level aggregation operators to enhance graph modeling capabilities. To further leverage the power of multi-agent in RL, AgentNet~\cite{martinkus2023agentbased} trains neural agents to walk through the graph and collectively decide on the output without applying traditional message passing. GAgN~\cite{liu2025graph} treats each node as an agent and proposes a decentralized graph structure agent network learning to infer global perception, enabling resistance to attacks on graph operators.

\textbf{LLM Agents}. 
With the powerful knowledge of LLMs, many works are exploring how to design and tune LLM agents to fully leverage the capabilities of LLMs in graph-modeling tasks. The core of these works is to post-train LLMs on graph data. Thus, LLM agents can automatically understand graph data through text-described graph information and carry out downstream tasks like node classification.
GL-Agent~\cite{wei2023versatile} and GraphGPT~\cite{tang2024graphgpt} initially propose LLM frameworks for agentic graph learning on various graphs.
GA~\cite{wang2023graph} uses LLM as the foundation model of an agent, and employs a memory mechanism to store nodes or edges embeddings for retrieval augmentation. 
Additionally, OFA~\cite{liu2024one} and ZeroG~\cite{li2024zerog} use the LLM agent to map nodes from different graphs into a unified semantic space to ensure the generalization ability across datasets and graph understanding tasks.
LLaGA~\cite{chen2024llaga} further transforms graph inputs into a token space comprehensible to LLMs. This transformation allows the use of the inherent reasoning abilities of LLMs for graph-related tasks without modifying the LLM parameters.
GraphInstruct~\cite{luo2024graphinstruct} constructs diverse graph generation processes and descriptions, and proposes a step-by-step masking training strategy for LLM supervised fine-tuning. Besides tuning to make models generate reasoning paths based on graph structure and text prompts, GraphWiz~\cite{chen2024graphwiz} further improves reasoning ability by using DPO~\cite{rafailov2023direct} to compare reasoning paths.

\textbf{LLM Multi-Agents}. 
Graph modeling can be seen as a complex task that involves multiple sub-modeling tasks, such as graph data processing, graph architecture selection, and graph task execution, which is difficult for a single agent to handle all. Therefore, recently, many works have introduced multiple LLMs to construct a multi-agent system, with each agent focusing on a specific sub-modeling task. For instance, GraphAgent-Reasoner~\cite{hu2024scalable} proposes the first multi-agent graph reasoning framework based on LLM without fine-tuning. The main LLM extracts the node and edge information from the text description and builds an agent for each node. These agents communicate with neighborhoods and update their states based on LLM's instructions. GraphAgent~\cite{yang2024graphagent} defines graph generation agents, task planning agents, and task execution agents. These agents work together to integrate language models and graph models, aiming to discover complex relational information and data semantic dependencies. Inspired by human problem solving processes, GraphTeam~\cite{li2024graphteam} further designs a multi-agent framework with three functional groups: input-output regularization, external knowledge retrieval, and graph-based problem solving.

\section{Applications}\label{sec:app}
\subsection{Scientific Computing}
Graph learning combined with agent systems has shown significant potential in the field of scientific computing, particularly in areas such as automated scientific discovery and bioinformatics analysis~\cite{wang2023scientific,zhang2024comprehensive}. 

Deep integration can be divided into two main aspects. First, using graphs to connect scientific knowledge and enhance agents' scientific understanding. Second, applying graph learning to model inherent structured data in some disciplines, such as molecular and protein structures, and then align with agents.
For example, on the one hand, SciAgents~\cite{ghafarollahi2024sciagents} enables automatic discovery of interdisciplinary knowledge and hypothesis generation by integrating large-scale ontology knowledge graphs and LLM multi-agent systems. Thought Graph~\cite{hsu2024thought} combines the LLM agent's structured Tree-of-Thought understanding process with a genetic knowledge base, achieving successes in gene set analysis. 
On the other hand, LLaMo~\cite{park2024llamo} proposes multi-level graph projectors and structure instruction tuning of LLM agents, targeting the challenge of converting between molecular structures and language modalities.

\subsection{Embodied AI}
In embodied AI, agents must perceive their surrounding environments and execute appropriate actions~\cite{liu2024aligning}. Graph-based representations are powerful tools in this domain and can be roughly divided into two categories.
First, scene graphs provide structured high-level representations of environments, enhancing the scene understanding capabilities of embodied agents. For instance, Ravichandran et al.~\cite{ravichandran2022hierarchical} embed scene graphs into a latent feature space using GNNs, enabling agents to learn effective navigation policies via reinforcement learning. SayPlan~\cite{rana2023sayplan} leverages a scene graph to guide LLMs in semantic search and refines the initial plan based on feedback from a graph simulator. SG-Nav~\cite{yin2024sg} incrementally constructs a scene graph during environment exploration, providing LLMs with scene context to reason about the locations of the goals and to select the exploration frontiers. 
Second, knowledge graphs have been employed in embodied AI to support more informed decision-making. For instance, ~\cite{yang2024task} represents task-relevant knowledge, while~\cite{qi2024safety} encodes safety protocols and physical constraints based on knowledge graphs. Scene-MMKG\cite{song2024scene} combines prompt-based schema extraction with knowledge from multiple sources to construct a scene-specific knowledge graph. By incorporating structured knowledge representations, these approaches enhance the abilities of embodied agents in planning, manipulation, and navigation.

\subsection{Game AI}
Graph structures are gaining increasing attention as an effective tool for representation and modeling in game AI. An important reason is that games involve strategic interactions between players, as well as between player-controlled characters and the game environment~\cite{perez2019general}. For example, in the analysis of two-player games, response graph models not only capture non-trivial properties of games but also reveal profound connections between different types of games~\cite{biggar2023graph}.
In multi-agent gaming systems, game abstraction mechanisms can simplify the policy learning process and enhance the asymptotic performance of algorithms~\cite{liu2020multi}. Specifically, in multi-agent sports games, graph-based generative models show superior performance in predicting multi-agent trajectories and in distribution similarity metrics~\cite{yeh2019diverse}. Recently, with the widespread application of LLMs, text-based games have become more and more famous~\cite{hu2024survey,shi2025monte}. Representatively, GATA~\cite{adhikari2020learning} adopts dynamic belief graphs to better understand the state of the game and make decisions in text-based games, achieving performance improvements over traditional text-based models.

\subsection{Agentic Information Retrieval}
Future research in agentic information retrieval can be significantly advanced by integrating graph learning to support structured retrieval planning and adaptive reasoning.
Representative approaches such as Think-on-Graph~\cite{sun2024thinkongraph, ma2025think} and  Graph Chain-of-Thought~\cite{jin2024graph}, where an LLM operates as an agent that interactively explores a knowledge graph (KG) to retrieve relevant information and perform reasoning. This is implemented via an iterative prompt loop where the LLM generates the next query or “move” on the graph, and the retrieved node facts are fed back into the prompt.
In parallel, RL-based optimization is being explored in agentic retrieval and reasoning. Recent methods~\cite{jin2025search, song2025r1, chen2025learning} train LLM agents with reward signals that encourage effective search queries and iterative evidence gathering, showing that models can learn when to invoke external knowledge and even retry searches after failures. 
A promising future direction is to unify graph-based agentic planning and RL-driven learning into one loop. For example, an autonomous retrieval agent that leverages knowledge graph relationships for decision making while continuously learning to refine its search and reasoning policy, thereby achieving more robust, interpretable, and efficient information retrieval in complex tasks.

\subsection{Industrial and Automation Systems}
In the era of Industry 4.0 and the Internet of Things (IoT), the focus is on achieving efficient and automated production through intelligent technologies~\cite{dong2023graph}. Currently, agent systems have begun to be applied in warehousing management, manufacturing scheduling, and other fields, providing key technical support for Industry 4.0~\cite{huang2022graph,duan2024data}.
In these industrial systems, the key capabilities that an agent should possess are scalability, dynamic evolvability, and robustness.
When it comes to scalability, in industrial systems, an agent often involves a large search space and a large number of tool invocations. There is also a great deal of relation between agents. Graph organization and learning can obtain a sparse association graph to reduce non-informative interactions and achieve efficient performance~\cite{kang2022learning,liu2024toolnet,hu2024learning}.
Graph-based models can also be flexible in dynamic environments with dynamic graph evolution~\cite{shen2024content,gupta2025deep}.
Also, the robustness of the agent system can be enhanced by learning the inherent patterns and structures in the graph data to filter out abnormal or noisy information in the system~\cite{wu2021graph,li2024coordinated}.

\subsection{Human Society}

LLM-based agents have held significant potential in analyzing and simulating human social behaviors in recent years. Research has shown that agents driven by LLM can effectively model information dissemination, emotional and attitudinal changes, and complex network formation within social networks~\cite{gao2024simulating}. 
An important reason is that LLM agents can simulate key dynamics in real-world social networks~\cite{papachristou2024network}. These include preferential attachment, triadic closure, homophily, and community structure. In addition, LLM agents can adjust their behavioral strategies according to the characteristics of the network. Based on this foundational ability, many works have adopted LLM agents for social network and user behavior simulation~\cite{muric2022large,gao2023s3,ji2024gag,ferraro2024agent}. Further, some works have adopted LLM agents for human society analysis. For example, \cite{shu2024llm} performs personality analysis in online social networks with LLM agents. \cite{hu2025simulating} employs LLM agents to analyze the mechanisms of social rumors spreading.

\section{Challenges and Future Opportunities}\label{sec:future}
\begin{table*}[t]
  \centering
  \caption{A collection of representative benchmarks and open-source toolkits.}
  \resizebox{0.9\textwidth}{!}{
    \begin{tabular}{cl|l|l|r}
    \toprule
    \multicolumn{2}{l|}{Benchmarks and Toolkits} & Agent Type & Evaluation Task (s) & \multicolumn{1}{l}{Link} \\
    \midrule
    \multirow{11}[2]{*}{\rotatebox{90}{General}} 
          & EPyMARL~\cite{papoudakis2021benchmarking} & RL Agent & Multi-Agent Coordination & \href{https://github.com/uoe-agents/epymarl}{Github} \\
          & BenchMARL~\cite{bettini2024benchmarl} & RL Agent & Multi-Agent Coordination & \href{https://github.com/facebookresearch/BenchMARL}{Github} \\
          & Memory Gym~\cite{pleines2025memory} & RL Agent & Agent Memory & \href{https://github.com/MarcoMeter/endless-memory-gym}{Github} \\
          & API-Bank~\cite{li2023api} & LLM Agent & Tool Usage & \href{https://github.com/AlibabaResearch/DAMO-ConvAI/tree/main/api-bank}{Github} \\
          & PlanBench~\cite{valmeekam2023planbench} & LLM Agent & Task Reasoning, Task Decomposition & \href{https://github.com/karthikv792/LLMs-Planning}{Github} \\
          & AgentBench~\cite{liu2024agentbench} & LLM Agent & General & \href{https://github.com/THUDM/AgentBench}{Github} \\
          & ToolBench~\cite{qintoolllm} & LLM Agent & Tool Usage &  \href{https://github.com/OpenBMB/ToolBench}{Github} \\
          & GTA~\cite{wang2024gta}   & LLM Agent & Tool Usage &  \href{https://github.com/open-compass/GTA}{Github} \\
          & TravelPlanner~\cite{xie2024travelplanner} & LLM Agent & Task Reasoning, Task Decomposition, Tool Usage & \href{https://github.com/OSU-NLP-Group/TravelPlanner}{Github} \\
          & $\tau$-bench~\cite{yao2024tau} & LLM Agent & Tool Usage, Agent-Environment Interaction & \href{https://github.com/sierra-research/tau-bench}{Github} \\
          & TOOLSANDBOX~\cite{lu2024toolsandbox} & LLM Agent & Tool Usage, Agent-Environment Interaction & \href{https://github.com/apple/ToolSandbox}{Github} \\
          & REALM-Bench~\cite{geng2025realm} & LLM Agent & Task Reasoning, Task Decomposition, Multi-Agent Coordination & \href{https://github.com/genglongling/REALM-Bench}{Github} \\
    \midrule
    \multirow{7}[2]{*}{\rotatebox{90}{Graph-Related}} 
          & LangGraph & LLM Agent & Tool Usage, Multi-Agent Coordination & \href{https://github.com/langchain-ai/langgraph}{Github} \\
          & GLBench~\cite{li2024glbench} & LLM Agent & Graph Modeling & \href{https://github.com/NineAbyss/GLBench}{Github} \\
          & ProGraph~\cite{li2024can} & LLM Agent & Graph Modeling & \href{https://github.com/BUPT-GAMMA/ProGraph}{Github} \\
          & GraphArena~\cite{tang2025grapharena} & LLM Agent & Graph Modeling & \href{https://github.com/squareRoot3/GraphArena}{Github} \\
          & GPTSwarm~\cite{zhuge2024gptswarm} & LLM Agent & Task Reasoning, Tool Usage, Multi-Agent Coordination  & \href{https://github.com/metauto-ai/gptswarm}{Github} \\
          & MACNET~\cite{qian2025scaling} & LLM Agent & Multi-Agent Coordination & \href{https://github.com/OpenBMB/ChatDev/tree/macnet}{Github} \\
          & ToolLinkOS~\cite{lumer2025graph} & LLM Agent & Tool Usage & \href{https://github.com/EliasLumer/Graph-RAG-Tool-Fusion-ToolLinkOS}{Github} \\
          & FLORA-Bench~\cite{zhang2025gnns} & LLM Agent & Task Reasoning, Task Decomposition & \href{https://github.com/youngsoul0731/Flora-Bench}{Github} \\
    \bottomrule
    \end{tabular}%
    }
  \label{tab:benchmark}%
\end{table*}%

\subsection{Benchmarking Evaluation}
As illustrated in Table~\ref{tab:benchmark}, there are currently numerous representative benchmarks and open-source toolkits that have been proposed for evaluations in different agent tasks, especially LLM-based agents. Graph-related agent benchmarks have also begun to gain attention in recent years. For the challenge aspects, existing benchmarks still differ in task definitions or evaluation data, making unified evaluation challenging. Furthermore, as LLM foundation models and RL technologies integrate deeply, future benchmarks could incorporate both and include graph-based models as a part. Also, for emerging complex scenarios in which graph learning shows great potential, like multi-agent reasoning, memory, and collaboration in large-scale dynamic environments, there is currently a lack of graph-centric agent benchmarks.

\subsection{Graph Foundation Models for Agents} 
In traditional graph modeling tasks, representative methods like GCN~\cite{kipf2017semisupervised} and GAT~\cite{velivckovic2018graph} adopt message passing for knowledge extraction, providing solid foundations for graph modeling. Based on such a foundation, many further works and applications are proposed to accelerate the development of the field. 
For example, some methods have begun to explore unified and effective graph foundation models (GFMs) for handling multiple graph domains simultaneously~\cite{liu2025gfm_survey,mao2024position}.
For AI agents, there are various graph-based methods for different tasks. However, there is a lack of foundational graph operators for widespread use. Therefore, developing effective GFMs in agent-based functions is a promising direction.
Furthermore, given that agent scenarios often need to handle complex tasks, GFMs for agents can be designed from the perspectives of \textbf{Effectiveness, Explainability, and Scalability (EES)}. Specifically, while serving effectiveness, GFMs can improve the interpretability of agent decisions by using logical relationships between nodes based on graph advantage~\cite{ying2019gnnexplainer,yuan2022explainability} and improve system scalability by organizing information and operators in a structured way~\cite{liu2024toolnet}.

\subsection{Privacy and Security}
For AI agents, graph organization and learning primarily establish connections among entities, such as agent nodes and external tools, to enable efficient message sharing and transmission, which may bring security issues.
% and data security
(1) \textbf{Data Privacy}: A future direction lies in enabling message sharing and transmission between nodes involving private information while ensuring privacy~\cite{maliah2017collaborative,he2024emerged,li2023private}, such as communication between private-domain agents~\cite{calvaresi2020personal,standen2025adversarial}. For example, in a healthcare setting, patient data is highly sensitive. If the agent's operators or the interaction of multiple agents involves information passing, it is crucial to ensure that patient privacy is protected.
(2) \textbf{Attack Defense}: 
Multi-agent communication and agent-environment interaction, which naturally form a graph topology, are susceptible to adversarial attacks~\cite{tu2021adversarial,sun2022adversarial,standen2025adversarial}. Attackers can send indistinguishable adversarial messages to severely disrupt the agent system's performance and privacy data leakage~\cite{liao2024eia,standen2025adversarial}.
Some works have explored the impact of multi-agent topology on system security~\cite{yu2024netsafe,wang2025g}. Highly connected topologies may make the system more susceptible to attacks, as malicious information can spread quickly among agents. Future research should focus on designing more secure multi-agent coordination and agent-environment interaction strategies. It is necessary to develop topologies and mechanisms that can defend against adversarial attacks while maintaining communication effectiveness.

\subsection{Multimodal Agents}
The agent paradigm using LLMs as foundation models has achieved remarkable performance in executing tasks in the language space. Research on agents and LLMs is still constantly evolving. For example, there have been some breakthroughs in large vision language models~\cite{xu2024lvlm} or multimodal large language models~\cite{liang2024survey} recently. They can also serve as the foundation model for next-generation AI agents. Multimodal agents will surpass current systems centered on language or single modalities, evolving into unified entities capable of understanding and integrating information streams such as text, vision, and speech~\cite{durante2024agent,xie2024large}.
Such agents should make decisions and learn autonomously from heterogeneous data sources. In this evolution, graph learning can play a vital role as multimodal data are inherently interlinked with complex and dynamic relationships. Graph techniques can abstract and connect different modalities, seamlessly integrating and aligning modality entities from the external world, perceived objects, and environmental states~\cite{peng2024learning,ektefaie2023multimodal,ning2025graphmm}.

\subsection{Model Context Protocol}
MCP is an open standard to provide a unified, structured, and secure way for LLM agents to exchange data with the outside world.
The MCP protocol defines how context information is exchanged between agents and external data, providing a standardized means for seamless integration between agent applications and external data sources, tools, and services~\cite{hou2025model,yang2025survey}. 
Since structured interaction between agents and external data is a key insight of MCP, graph learning can enhance MCP from two promising perspectives. (1) \textbf{Efficient Data Integration}: Various data types and sources can be integrated into a unified KG. Graph learning can uncover hidden relationships and patterns, improving KG construction. This enriches the knowledge base accessible via the MCP protocol, providing AI models with more comprehensive and accurate context for better problem-solving. (2) \textbf{Personalized Recommendations}: By modeling agent behavior and preferences, graph learning can provide personalized tool and data source recommendations. The MCP protocol can be extended by using this to quickly locate and call the most suitable tools and data for different agents, enhancing their experience and productivity.

\subsection{Open Agent Network}
A single agent can hardly optimize security, latency, domain expertise, and tool usage all at once.
In the rapid development of agents empowered with LLM, agents are shifting from single, monolithic, general-purpose chatbots to ecosystems and swarms of specialized agents~\cite{guo2024large,tran2025multi}.
An Open Agent Network (OAN) would be a promising, public, and decentralized network where agents are registered and orchestrated~\cite{willmott2001agentcities}. It could be a unified, heterogeneous, and large-scale graph, including potential network participants like agents, tools, tasks (applications), humans, and policies (regulators) in the future. In the prospects for OAN, the network infrastructure offers discovery, routing, invocation, settlement, and reputation services.
Any compliant third party can publish or invoke agents, yielding a composable agent graph.
Within the OAN, graphs can weave together several complementary capabilities. (1) \textbf{OAN Effectiveness}: Graph learning can predict the optimal call path by incorporating latency and cost, continuously update the OAN graph online, and inductively generate representations for newly added nodes. (2) \textbf{OAN Risk Control}: Graph representations enable trust and risk analysis by applying tasks like node classification and link prediction to gauge the reputation of newly registered agents and highlight fragile dependency chains, while temporal patterns extracted from edge logs can be adopted to detect potential anomalies.

\section{Conclusion}\label{sec:conclusion}
In this paper, we provide a comprehensive and systematic review of the intersection of graph techniques and AI agents, with graph-empowered agent functionalities, including planning, execution, memory, and multi-agent coordination. In addition, we also investigate how AI agent paradigms can enhance graph learning, such as graph annotation, synthesis, and understanding. Based on the detailed review, we then summarize meaningful applications, open questions, and future opportunities for this rapidly developing intersection field, offering new potential approaches for next-generation agents facing increasingly complex and messy task information. Related resources are organized and updated in \href{https://github.com/YuanchenBei/Awesome-Graphs-Meet-Agents}{\textbf{Github link}} for the research and industrial communities.

\bibliographystyle{IEEEtran}
\bibliography{custom}

\vfill

\end{document}